\documentclass[11pt, a4paper, logo, onecolumn, copyright, nonumbering]{deepmind}

\usepackage[authoryear, sort&compress, round]{natbib}
\usepackage{amsmath,amsfonts,bm}

\def\eqref#1{equation~\ref{#1}}

\def\1{\bm{1}}

\DeclareMathAlphabet{\mathsfit}{\encodingdefault}{\sfdefault}{m}{sl}
\SetMathAlphabet{\mathsfit}{bold}{\encodingdefault}{\sfdefault}{bx}{n}

\usepackage{floatrow}
\usepackage{varwidth}
\usepackage[]{algorithm2e}
\usepackage{algpseudocode}
\usepackage{bbm}
\usepackage{wrapfig}
\usepackage{siunitx}
\usepackage{cleveref}

\usepackage{fancyvrb}
\usepackage{graphicx}
\usepackage{xcolor}

\usepackage[bottom]{footmisc}

\makeatletter
\def\PYG@reset{\let\PYG@it=\relax \let\PYG@bf=\relax%
    \let\PYG@ul=\relax \let\PYG@tc=\relax%
    \let\PYG@bc=\relax \let\PYG@ff=\relax}
\def\PYG@tok#1{\csname PYG@tok@#1\endcsname}
\def\PYG@toks#1+{\ifx\relax#1\empty\else%
    \PYG@tok{#1}\expandafter\PYG@toks\fi}
\def\PYG@do#1{\PYG@bc{\PYG@tc{\PYG@ul{%
    \PYG@it{\PYG@bf{\PYG@ff{#1}}}}}}}
\def\PYG#1#2{\PYG@reset\PYG@toks#1+\relax+\PYG@do{#2}}

\@namedef{PYG@tok@w}{\def\PYG@tc##1{\textcolor[rgb]{0.73,0.73,0.73}{##1}}}
\@namedef{PYG@tok@c}{\let\PYG@it=\textit\def\PYG@tc##1{\textcolor[rgb]{0.24,0.48,0.48}{##1}}}
\@namedef{PYG@tok@cp}{\def\PYG@tc##1{\textcolor[rgb]{0.61,0.40,0.00}{##1}}}
\@namedef{PYG@tok@k}{\let\PYG@bf=\textbf\def\PYG@tc##1{\textcolor[rgb]{0.00,0.50,0.00}{##1}}}
\@namedef{PYG@tok@kp}{\def\PYG@tc##1{\textcolor[rgb]{0.00,0.50,0.00}{##1}}}
\@namedef{PYG@tok@kt}{\def\PYG@tc##1{\textcolor[rgb]{0.69,0.00,0.25}{##1}}}
\@namedef{PYG@tok@o}{\def\PYG@tc##1{\textcolor[rgb]{0.40,0.40,0.40}{##1}}}
\@namedef{PYG@tok@ow}{\let\PYG@bf=\textbf\def\PYG@tc##1{\textcolor[rgb]{0.67,0.13,1.00}{##1}}}
\@namedef{PYG@tok@nb}{\def\PYG@tc##1{\textcolor[rgb]{0.00,0.50,0.00}{##1}}}
\@namedef{PYG@tok@nf}{\def\PYG@tc##1{\textcolor[rgb]{0.00,0.00,1.00}{##1}}}
\@namedef{PYG@tok@nc}{\let\PYG@bf=\textbf\def\PYG@tc##1{\textcolor[rgb]{0.00,0.00,1.00}{##1}}}
\@namedef{PYG@tok@nn}{\let\PYG@bf=\textbf\def\PYG@tc##1{\textcolor[rgb]{0.00,0.00,1.00}{##1}}}
\@namedef{PYG@tok@ne}{\let\PYG@bf=\textbf\def\PYG@tc##1{\textcolor[rgb]{0.80,0.25,0.22}{##1}}}
\@namedef{PYG@tok@nv}{\def\PYG@tc##1{\textcolor[rgb]{0.10,0.09,0.49}{##1}}}
\@namedef{PYG@tok@no}{\def\PYG@tc##1{\textcolor[rgb]{0.53,0.00,0.00}{##1}}}
\@namedef{PYG@tok@nl}{\def\PYG@tc##1{\textcolor[rgb]{0.46,0.46,0.00}{##1}}}
\@namedef{PYG@tok@ni}{\let\PYG@bf=\textbf\def\PYG@tc##1{\textcolor[rgb]{0.44,0.44,0.44}{##1}}}
\@namedef{PYG@tok@na}{\def\PYG@tc##1{\textcolor[rgb]{0.41,0.47,0.13}{##1}}}
\@namedef{PYG@tok@nt}{\let\PYG@bf=\textbf\def\PYG@tc##1{\textcolor[rgb]{0.00,0.50,0.00}{##1}}}
\@namedef{PYG@tok@nd}{\def\PYG@tc##1{\textcolor[rgb]{0.67,0.13,1.00}{##1}}}
\@namedef{PYG@tok@s}{\def\PYG@tc##1{\textcolor[rgb]{0.73,0.13,0.13}{##1}}}
\@namedef{PYG@tok@sd}{\let\PYG@it=\textit\def\PYG@tc##1{\textcolor[rgb]{0.73,0.13,0.13}{##1}}}
\@namedef{PYG@tok@si}{\let\PYG@bf=\textbf\def\PYG@tc##1{\textcolor[rgb]{0.64,0.35,0.47}{##1}}}
\@namedef{PYG@tok@se}{\let\PYG@bf=\textbf\def\PYG@tc##1{\textcolor[rgb]{0.67,0.36,0.12}{##1}}}
\@namedef{PYG@tok@sr}{\def\PYG@tc##1{\textcolor[rgb]{0.64,0.35,0.47}{##1}}}
\@namedef{PYG@tok@ss}{\def\PYG@tc##1{\textcolor[rgb]{0.10,0.09,0.49}{##1}}}
\@namedef{PYG@tok@sx}{\def\PYG@tc##1{\textcolor[rgb]{0.00,0.50,0.00}{##1}}}
\@namedef{PYG@tok@m}{\def\PYG@tc##1{\textcolor[rgb]{0.40,0.40,0.40}{##1}}}
\@namedef{PYG@tok@gh}{\let\PYG@bf=\textbf\def\PYG@tc##1{\textcolor[rgb]{0.00,0.00,0.50}{##1}}}
\@namedef{PYG@tok@gu}{\let\PYG@bf=\textbf\def\PYG@tc##1{\textcolor[rgb]{0.50,0.00,0.50}{##1}}}
\@namedef{PYG@tok@gd}{\def\PYG@tc##1{\textcolor[rgb]{0.63,0.00,0.00}{##1}}}
\@namedef{PYG@tok@gi}{\def\PYG@tc##1{\textcolor[rgb]{0.00,0.52,0.00}{##1}}}
\@namedef{PYG@tok@gr}{\def\PYG@tc##1{\textcolor[rgb]{0.89,0.00,0.00}{##1}}}
\@namedef{PYG@tok@ge}{\let\PYG@it=\textit}
\@namedef{PYG@tok@gs}{\let\PYG@bf=\textbf}
\@namedef{PYG@tok@gp}{\let\PYG@bf=\textbf\def\PYG@tc##1{\textcolor[rgb]{0.00,0.00,0.50}{##1}}}
\@namedef{PYG@tok@go}{\def\PYG@tc##1{\textcolor[rgb]{0.44,0.44,0.44}{##1}}}
\@namedef{PYG@tok@gt}{\def\PYG@tc##1{\textcolor[rgb]{0.00,0.27,0.87}{##1}}}
\@namedef{PYG@tok@err}{\def\PYG@bc##1{{\setlength{\fboxsep}{\string -\fboxrule}\fcolorbox[rgb]{1.00,0.00,0.00}{1,1,1}{\strut ##1}}}}
\@namedef{PYG@tok@kc}{\let\PYG@bf=\textbf\def\PYG@tc##1{\textcolor[rgb]{0.00,0.50,0.00}{##1}}}
\@namedef{PYG@tok@kd}{\let\PYG@bf=\textbf\def\PYG@tc##1{\textcolor[rgb]{0.00,0.50,0.00}{##1}}}
\@namedef{PYG@tok@kn}{\let\PYG@bf=\textbf\def\PYG@tc##1{\textcolor[rgb]{0.00,0.50,0.00}{##1}}}
\@namedef{PYG@tok@kr}{\let\PYG@bf=\textbf\def\PYG@tc##1{\textcolor[rgb]{0.00,0.50,0.00}{##1}}}
\@namedef{PYG@tok@bp}{\def\PYG@tc##1{\textcolor[rgb]{0.00,0.50,0.00}{##1}}}
\@namedef{PYG@tok@fm}{\def\PYG@tc##1{\textcolor[rgb]{0.00,0.00,1.00}{##1}}}
\@namedef{PYG@tok@vc}{\def\PYG@tc##1{\textcolor[rgb]{0.10,0.09,0.49}{##1}}}
\@namedef{PYG@tok@vg}{\def\PYG@tc##1{\textcolor[rgb]{0.10,0.09,0.49}{##1}}}
\@namedef{PYG@tok@vi}{\def\PYG@tc##1{\textcolor[rgb]{0.10,0.09,0.49}{##1}}}
\@namedef{PYG@tok@vm}{\def\PYG@tc##1{\textcolor[rgb]{0.10,0.09,0.49}{##1}}}
\@namedef{PYG@tok@sa}{\def\PYG@tc##1{\textcolor[rgb]{0.73,0.13,0.13}{##1}}}
\@namedef{PYG@tok@sb}{\def\PYG@tc##1{\textcolor[rgb]{0.73,0.13,0.13}{##1}}}
\@namedef{PYG@tok@sc}{\def\PYG@tc##1{\textcolor[rgb]{0.73,0.13,0.13}{##1}}}
\@namedef{PYG@tok@dl}{\def\PYG@tc##1{\textcolor[rgb]{0.73,0.13,0.13}{##1}}}
\@namedef{PYG@tok@s2}{\def\PYG@tc##1{\textcolor[rgb]{0.73,0.13,0.13}{##1}}}
\@namedef{PYG@tok@sh}{\def\PYG@tc##1{\textcolor[rgb]{0.73,0.13,0.13}{##1}}}
\@namedef{PYG@tok@s1}{\def\PYG@tc##1{\textcolor[rgb]{0.73,0.13,0.13}{##1}}}
\@namedef{PYG@tok@mb}{\def\PYG@tc##1{\textcolor[rgb]{0.40,0.40,0.40}{##1}}}
\@namedef{PYG@tok@mf}{\def\PYG@tc##1{\textcolor[rgb]{0.40,0.40,0.40}{##1}}}
\@namedef{PYG@tok@mh}{\def\PYG@tc##1{\textcolor[rgb]{0.40,0.40,0.40}{##1}}}
\@namedef{PYG@tok@mi}{\def\PYG@tc##1{\textcolor[rgb]{0.40,0.40,0.40}{##1}}}
\@namedef{PYG@tok@il}{\def\PYG@tc##1{\textcolor[rgb]{0.40,0.40,0.40}{##1}}}
\@namedef{PYG@tok@mo}{\def\PYG@tc##1{\textcolor[rgb]{0.40,0.40,0.40}{##1}}}
\@namedef{PYG@tok@ch}{\let\PYG@it=\textit\def\PYG@tc##1{\textcolor[rgb]{0.24,0.48,0.48}{##1}}}
\@namedef{PYG@tok@cm}{\let\PYG@it=\textit\def\PYG@tc##1{\textcolor[rgb]{0.24,0.48,0.48}{##1}}}
\@namedef{PYG@tok@cpf}{\let\PYG@it=\textit\def\PYG@tc##1{\textcolor[rgb]{0.24,0.48,0.48}{##1}}}
\@namedef{PYG@tok@c1}{\let\PYG@it=\textit\def\PYG@tc##1{\textcolor[rgb]{0.24,0.48,0.48}{##1}}}
\@namedef{PYG@tok@cs}{\let\PYG@it=\textit\def\PYG@tc##1{\textcolor[rgb]{0.24,0.48,0.48}{##1}}}

\def\PYGZus{\char`\_}

\def\PYGZhy{\char`\-}

\makeatother

\title{Jasmine: A Simple, Performant and Scalable JAX-based World Modeling Codebase}%
\correspondingauthor{[mihir,alfred,franz]@pdoom.org}


\reportnumber{001} %

\author{
{\Authfont
Mihir Mahajan\textsuperscript{*1,2},
Alfred Nguyen\textsuperscript{*1,2},
Franz Srambical\textsuperscript{*1,2}}
\leavevmode \protect\\[1em]
{\Authfont
Stefan Bauer\textsuperscript{2}}
\leavevmode \protect\\[1em]
{\Affilfont
\textsuperscript{*}Contributed equally\protect\\
\textsuperscript{1}p(doom), \textsuperscript{2}TUM 
}}

\begin{abstract}
    While world models are increasingly positioned as a pathway to overcoming data scarcity in domains such as robotics, open training infrastructure for world modeling remains nascent. We introduce Jasmine, a performant JAX-based world modeling codebase that scales from single hosts to hundreds of accelerators with minimal code changes. Jasmine achieves an order-of-magnitude faster reproduction of the CoinRun case study compared to prior open implementations, enabled by performance optimizations across data loading, training and checkpointing. The codebase guarantees fully reproducible training and supports diverse sharding configurations. By pairing Jasmine with curated large-scale datasets, we establish infrastructure for rigorous benchmarking pipelines across model families and architectural ablations.
\end{abstract}

\begin{document}
\maketitle

\section{Introduction}

Over the past decades, the field of deep learning has increasingly been shaped by methods that leverage vast data troves \citep{jozefowicz2016exploring, radford2018improving, radford2019language, chowdhery2022palm, raffel2020exploring, JMLR:v23:21-0998, deng2009imagenet}, and paradigms that unlock new ones \citep{guo2025deepseek, christiano2017deep, radford2021learning, srambical2025crowd-sourcing, silver2016mastering, berner2019dota}. Internet-scale pre-training, preference modeling, and reinforcement learning using verification signals offer a compelling pathway for language models to attain human-level performance \citep{LuongLockhart2025GeminiIMO, LinCheng2025GeminiICPC}, yet data is increasingly bottlenecking progress from spiky towards general intelligence. While some domains can leverage user feedback from deployed products for iterative model improvement \citep{cursor2025tab}, domains like robotics cannot afford such a privilege.

One paradigm proposed by the research community to overcome the data scarcity in those domains is that of world models \citep{ha2018world}. World models can aid frontier model development in numerous ways, but one particularly ambitious goal of the community is to train a world model to act as a simulation of the real world \citep{bruce2024genie, parker2022evolving, deepmind2025genie3, agarwal2025cosmos}, in order to train agents in that simulation \citep{hafner2025training}, via an adaptive curriculum \citep{parker2022evolving}, or otherwise. This regime requires the compounding error of the world model to be orders of magnitude smaller than when solely used for short-term look-ahead. The feasibility of such a world model in its truest sense is entirely understudied.
Jasmine provides the foundational infrastructure for future empirical investigation of how compute and data requirements scale with environment complexity for downstream agent training.

\begin{figure}
  \centering
  \includegraphics[width=\textwidth, trim=1 0 0 0, clip]{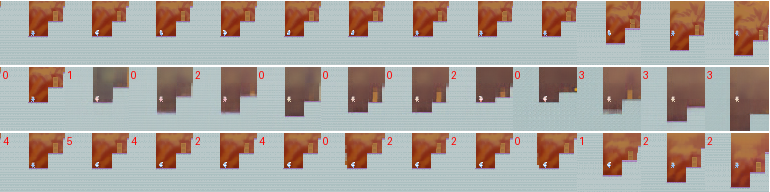}
  \caption{Autoregressive sampling of Jafar \citep{willi2024jafar} (middle row) and Jasmine (bottom row) on the CoinRun case study with four conditioning frames (conditioning frames not shown). 
  The top row shows the ground-truth sequence. 
}

\label{fig:sampling}
\end{figure}

\section{Jasmine}
\label{sec:jasmine}

\textbf{Our contributions} in this work are threefold: i) we introduce Jasmine, a highly optimized and scalable JAX-based codebase for world modeling, which we use to reproduce Genie's CoinRun case study \citep{bruce2024genie,cobbe2019leveraging} an order of magnitude faster than prior work \citep{willi2024jafar}. This speedup is the result of infrastructure optimizations, including a fully reproducible, scalable training and data pipeline built on the JAX \citep{jax2018github} ecosystem. ii) We find that a critical modification to the original Genie architecture, prepending latent actions instead of adding them to video embeddings, is required for the world model to yield generations that faithfully reproduce the CoinRun environment. iii) Finally, we openly release the Jasmine codebase, along with pretrained checkpoints, curated datasets, model inspection notebooks, and a dataset of dense IDE interactions captured during Jasmine's development, providing the first openly published dataset of months-long software engineering.

Jasmine implements the Genie \citep{bruce2024genie} architecture, enabling training of interactive environments from unlabeled videos. The architecture includes a video tokenizer, which encodes videos into tokens, a latent action model (LAM) that extracts latent actions between video frames, and a dynamics model that predicts the tokens of the next frame based on the previous tokens and corresponding latent actions. At sampling time, the LAM is discarded and replaced by input from the user. All modules use an ST-Transformer \citep{ho2019axial} backbone which approximates full attention by performing intra-frame (spatial) followed by inter-frame (temporal) attention, thus reducing the attention sequence length.
The tokenizer uses a VQ-VAE \citep{van2017neural} to encode image patches using reconstruction, vector-quantization, and commitment losses.
To train an action-conditioned video-generation model from unlabeled videos, Genie learns latent actions \citep{schmidt2023learning}. Like the tokenizer, the LAM uses a VQ-VAE, with its codebook representing the latent actions. The model learns to distill information from future frames into this bottlenecked codebook:
Frames $x_{0:t}$ are encoded, producing latent actions $a_{0:t}$, which the decoder receives along with past frames $x_{0:t-1}$ to predict the next frame $x_{t}$. A temporal causal mask allows the entire sequence to be processed in a single forward pass.
The dynamics model is a decoder-only transformer that predicts future frames conditioned on past frames and corresponding latent actions. Genie uses MaskGIT \citep{chang2022maskgit}, which masks input-tokens at training time, similar in spirit to BERT \citep{devlin2019bert}. Unlike MaskGIT, Genie masks with probability $p \sim U(0.5, 1)$ (refer to Appendix \ref{sec:maskgit_to_videos} for details about extending MaskGIT to videos).

\begin{figure}[htb]
  \centering
  \includegraphics[width=\textwidth]{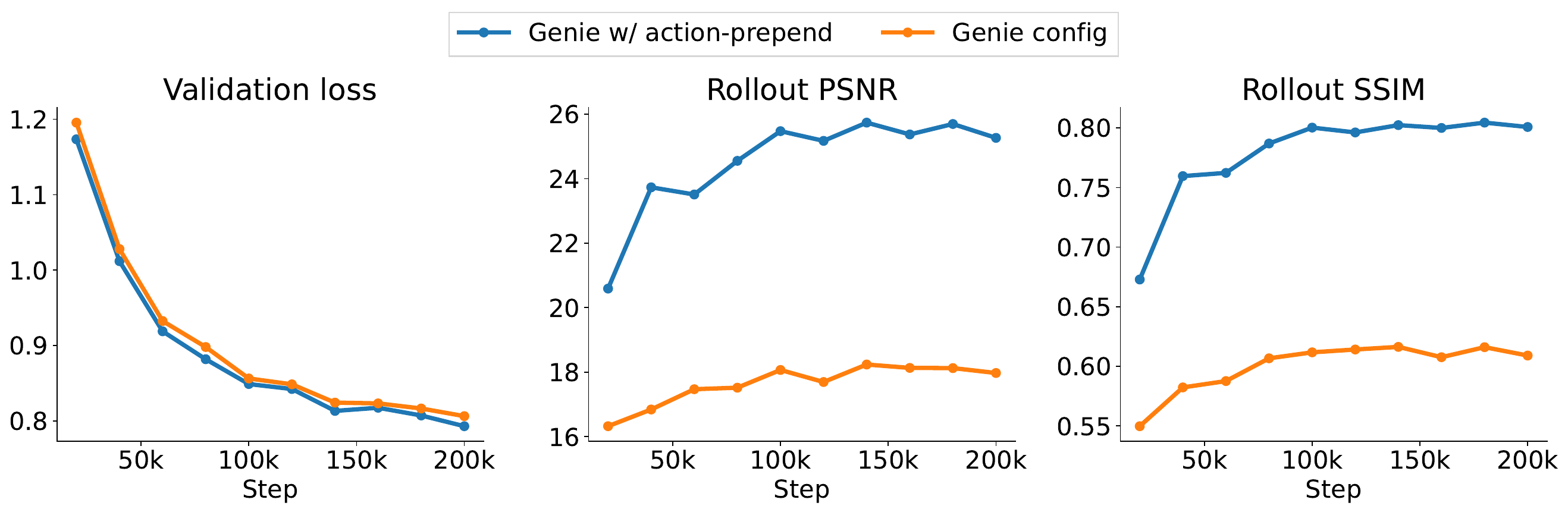}
  \caption{Validation metrics of the CoinRun case study (patch size 4). While the loss (left) is similar between the default Genie configuration and our minimal modification, rollout metrics (middle and right, refer to \Cref{sec:experiment-metrics}) differ substantially.}
  \label{fig:case-study-plots}
\end{figure}

Building upon prior work \citep{willi2024jafar} that openly published a reimplementation of the Genie \citep{bruce2024genie} architecture, we release a highly optimized JAX-based world modeling codebase amenable to scale. Jasmine implements a range of baselines, including MaskGIT-based \citep{chang2022maskgit}, fully causal \citep{srambical2024going}, and diffusion-based approaches (Appendix \ref{sec:diffusion-baseline}). The codebase depends solely  on battle-tested libraries from the Google ecosystem (JAX, NNX, Grain \citep{grain2023github}, Orbax, Optax, Treescope \citep{johnson2024penzai}, ArrayRecord \citep{ArrayRecord}), and scales from single hosts to hundreds of accelerators using XLA.
Jasmine supports complex sharding configurations in a few lines of code through Shardy \citep{openxla-shardy}. It provides asynchronous distributed checkpointing with configurable policies, process-parallel dataloading, and checkpointing of model, optimizer, and data loader states. Training runs are bitwise deterministic, yielding identical loss curves under identical seeds (Appendix \ref{sec:bitwise_deterministic}).
To enable efficient large-scale experimentation, Jasmine integrates mixed-precision, FlashAttention via cuDNN SDPA \citep{NVIDIA_cuDNN_Attention}, activation checkpointing, host memory offloading, and index-shuffling during data loading. The codebase follows the shape suffix convention of \citet{shazeer2024shape}, aiming to provide a didactic implementation of modern world modeling architectures.

We run the reproducible case study described in \citet{bruce2024genie} by generating a dataset containing 50M transitions of CoinRun, an environment of the Procgen benchmark \citep{cobbe2019leveraging} (Appendix \ref{sec:coinrun-case-study}). In contrast to \citet{bruce2024genie} we find that strict adoption of the architecture and hyperparameters described in their case study leads to deteriorating autoregressive generations in both Jasmine and Jafar \citep{willi2024jafar} (Figure \ref{fig:sampling_prepend_vs_no_prepend}, middle row). However, a minimal modification of the case study setting, namely prepending latent actions instead of adding them to the video embeddings, yields autoregressive generations that faithfully simulate the CoinRun environment (Figure \ref{fig:sampling_prepend_vs_no_prepend}, bottom row and Figure \ref{fig:case-study-plots}). We hypothesize that this discrepancy between \citet{bruce2024genie} and our work stems from an ambiguity in extending MaskGIT to videos (refer to Appendix \ref{sec:maskgit_to_videos} for further discussion).

\begin{figure}[htb]
  \centering
  \includegraphics[width=0.75\textwidth]{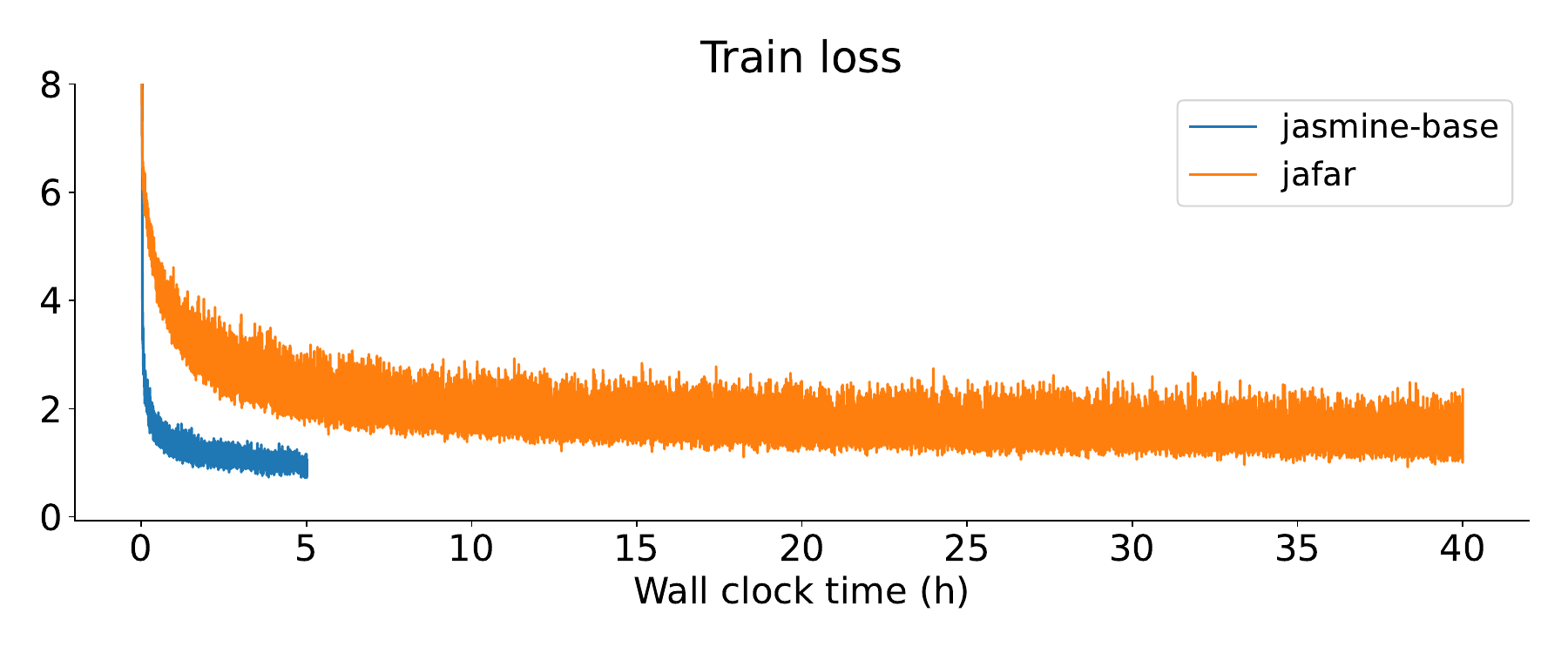}
  \caption{An order of magnitude faster convergence in wall-clock time in Jasmine (blue) compared to Jafar \citep{willi2024jafar} (orange). We report the train loss since Jafar does not collect validation metrics. Refer to Appendix \ref{sec:arch-ablations} for Jasmine's validation metrics. Jasmine's lower variance stems from a subtle refinement in its batched masking logic (Appendix \ref{sec:jafar-batched-masking}).}
  \label{fig:wc_time}
\end{figure}

Beyond openly publishing\footnote{\url{https://github.com/p-doom/jasmine}} the Jasmine codebase, we recorded every keystroke during Jasmine's development using \emph{crowd-code} \citep{srambical2025crowd-sourcing}, a VS Code/Cursor extension that enables crowd-sourcing dense IDE interaction data. To our knowledge, this represents one of the first open datasets capturing the full temporal scale of months-long software development, thus laying groundwork for future work in behaviour cloning, goal-conditioning, and verification signal mining.

\section{Experiments}

We evaluate the performance of Jasmine and analyze the impact of its core components through rigorous ablations. Using Jasmine, we reproduce the CoinRun case study at a patch size of 16 in under nine hours on a single GPU, compared to over 100 hours reported in prior work \citep{willi2024jafar} under the same setting (Figure \ref{fig:wc_time}). We present ablations identifying the factors responsible for this speedup in Table \ref{tab:infra-ablations}.
\Cref{sec:arch-ablations} further reports results from architectural modifications including replacing the latent action model with ground-truth actions, ablating co-training, adopting fully causal and diffusion baselines, and setting the feedforward expansion factor to four.

\paragraph{Architectural optimizations}
\label{sec:speed-arch-ablations}
We adapt Genie's architectural choices by integrating best practices from the language modeling community. Specifically, we use a feedforward expansion factor of four relative to the model dimension, following common practice in large-scale language modeling \citep{raffel2020exploring,radford2019language,brown2020language}. We simultaneously reduce network depth, resulting in lower overall parameter count than the Genie defaults, thus achieving higher throughput (Table \ref{tab:ffn-ablation}) while maintaing competitive performance (Figure \ref{fig:arch-ablations}). We employ the warmup-stable-decay (WSD) learning rate schedule \citep{zhai2022scaling, mahajan2018exploring}, which allows flexible training durations by resuming from a checkpoint prior to the decay phase. Unlike \citep{bruce2024genie}, we warm up from and decay the learning rate to zero, in line with established best practices \citep{zhai2022scaling}.
We further compare co-training LAM and dynamics model (as done in \citet{bruce2024genie}) with pre-training the LAM (as done in \citet{willi2024jafar}), embedding ground-truth actions instead of using the latent action model (Appendix \ref{sec:ablation-gt}), and replacing MaskGIT with fully causal and diffusion baselines. Co-training, pre-training the LAM, and using ground-truth actions are all competitive (Figure \ref{fig:arch-ablations}), while the fully causal baseline underperforms in the 200k steps training regime (\Cref{fig:causal-ablations}). However, our results indicate that the fully causal baseline in particular may benefit from longer training. Diffusion-forcing \citep{chen2024diffusion} outperforms MaskGIT, even when using identical per-frame sampling step counts and untuned hyperparameters (Figure \ref{fig:diffusion-plot}, Appendix \ref{sec:diffusion-baseline}).

\begin{table}[htb]
\begin{center}
\begin{tabular}{lcc}
\toprule
 & \textbf{Throughput (bs=36)} & \textbf{Throughput (bs=2048)} \\
\midrule
Jasmine-base & 1.00x	            & 1.00x \\
1x feedforward expansion & 0.93x	            & 0.79x \\
\bottomrule
\end{tabular}
\caption{Training throughput with a feedforward expansion factor of one, relative to Jasmine-base. We double the number of layers compared to Jasmine-base to roughly match the parameter count (refer to Genie's default configuration in \ref{tab:hparams}).}
\label{tab:ffn-ablation}
\end{center}
\end{table}

\paragraph{Infrastructure optimizations}
\label{sec:speed-infra-ablations}
A substantial portion of our speedup compared to \citet{willi2024jafar} arises from our data loader design (Tables \ref{tab:chunking-formats-1} and \ref{tab:chunking-formats-2}). We use Grain for data loading with prefetching enabled and preprocess datasets into ArrayRecords \citep{ArrayRecord}, a file format optimized for random access indexing. The chosen chunking strategy significantly affects throughput, and we describe our configuration in Appendix \ref{sec:throughput-ablation}. Jasmine further leverages FlashAttention \citep{dao2022flashattention} via cuDNN SDPA \citep{NVIDIA_cuDNN_Attention} and mixed precision training with bfloat16. We report throughputs when ablating mixed precision, FlashAttention and Grain in Table \ref{tab:infra-ablations}.

\section{Related Work and Discussion}
Although research on world models with its inception decades ago \citep{sutton1991dyna} has matured over the years \citep{ha2018world,hafner2019learning,hafner2019dream,hafner2020mastering,hafner2023mastering,alonso2024diffusion}, they have only been scaled up recently \citep{hafner2025training,bruce2024genie,parkerholder2024genie2,deepmind2025genie3,valevski2024diffusion,agarwal2025cosmos,hu2023gaia,guo2025mineworld,decart2024oasis,lucid2024lucidv1,li2025hunyuan,pearce2024scaling}. While the open training ecosystem in language modeling provides mature solutions for large-scale language pretraining \citep{megatron-lm,maxtext-library}, open training infrastructure for world modeling is still nascent \citep{wayfarer_labs_owl_wms,Savov_2025_CVPR}. Closest to our work is \citet{willi2024jafar}, an open-source reproduction of Genie \citep{bruce2024genie}, which we build upon and significantly extend.

With Jasmine we make progress towards democratizing world modeling research. Alongside the codebase, we openly release checkpoints and datasets for CoinRun, Atari and Doom, as well as dense IDE interaction data collected over months of research engineering\footnote{Datasets published under the CC0 license at \url{https://huggingface.co/datasets/p-doom}}. While Jasmine greatly accelerates wall-clock convergence compared to prior work, it has yet to match throughput efficiencies of frontier language model implementations.

\subsubsection*{Acknowledgments}
We thank Matthew T. Jackson, Andrea Dittadi and Diego Marti Monso for useful discussions as well as the Jafar authors for openly publishing their repository. 
The authors gratefully acknowledge the computing time provided on the high-performance computer HoreKa by the National High-Performance Computing Center at KIT (NHR@KIT). This center is jointly supported by the Federal Ministry of Education and Research and the Ministry of Science, Research and the Arts of Baden-Württemberg, as part of the National High-Performance Computing (NHR) joint funding program (https://www.nhr-verein.de/en/our-partners). HoreKa is partly funded by the German Research Foundation (DFG).

\bibliography{bibliography}

\begin{thebibliography}{65}
\providecommand{\natexlab}[1]{#1}
\providecommand{\url}[1]{\texttt{#1}}
\expandafter\ifx\csname urlstyle\endcsname\relax
  \providecommand{\doi}[1]{doi: #1}\else
  \providecommand{\doi}{doi: \begingroup \urlstyle{rm}\Url}\fi

\bibitem[Agarwal et~al.(2025)Agarwal, Ali, Bala, Balaji, Barker, Cai,
  Chattopadhyay, Chen, Cui, Ding, et~al.]{agarwal2025cosmos}
N.~Agarwal, A.~Ali, M.~Bala, Y.~Balaji, E.~Barker, T.~Cai, P.~Chattopadhyay,
  Y.~Chen, Y.~Cui, Y.~Ding, et~al.
\newblock Cosmos world foundation model platform for physical ai.
\newblock \emph{arXiv preprint arXiv:2501.03575}, 2025.

\bibitem[Alonso et~al.(2024)Alonso, Jelley, Micheli, Kanervisto, Storkey,
  Pearce, and Fleuret]{alonso2024diffusion}
E.~Alonso, A.~Jelley, V.~Micheli, A.~Kanervisto, A.~J. Storkey, T.~Pearce, and
  F.~Fleuret.
\newblock Diffusion for world modeling: Visual details matter in atari.
\newblock In \emph{Advances in neural information processing systems},
  volume~37, 2024.

\bibitem[Ball et~al.(2025)Ball, Bauer, Belletti, Brownfield, Ephrat, Fruchter,
  Gupta, Holsheimer, Holynski, Hron, Kaplanis, Limont, McGill, Oliveira,
  Parker-Holder, Perbet, Scully, Shar, Spencer, Tov, Villegas, Wang, Yung,
  Baetu, Berbel, Bridson, Bruce, Buttimore, Chakera, Chandra, Collins, Cullum,
  Damoc, Dasagi, Gazeau, Gbadamosi, Han, Hirst, Kachra, Kerley, Kjems,
  Knoepfel, Koriakin, Lo, Lu, Mehring, Moufarek, Nandwani, Oliveira, Pardo,
  Park, Pierson, Poole, Ran, Salimans, Sanchez, Saprykin, Shen, Sidhwani,
  Smith, Stanton, Tomlinson, Vijaykumar, Wang, Wingfield, Wong, Xu, Yew, Young,
  Zubov, Eck, Erhan, Kavukcuoglu, Hassabis, Gharamani, Hadsell, van~den Oord,
  Mosseri, Bolton, Singh, and Rockt{\"a}schel]{deepmind2025genie3}
P.~J. Ball, J.~Bauer, F.~Belletti, B.~Brownfield, A.~Ephrat, S.~Fruchter,
  A.~Gupta, K.~Holsheimer, A.~Holynski, J.~Hron, C.~Kaplanis, M.~Limont,
  M.~McGill, Y.~Oliveira, J.~Parker-Holder, F.~Perbet, G.~Scully, J.~Shar,
  S.~Spencer, O.~Tov, R.~Villegas, E.~Wang, J.~Yung, C.~Baetu, J.~Berbel,
  D.~Bridson, J.~Bruce, G.~Buttimore, S.~Chakera, B.~Chandra, P.~Collins,
  A.~Cullum, B.~Damoc, V.~Dasagi, M.~Gazeau, C.~Gbadamosi, W.~Han, E.~Hirst,
  A.~Kachra, L.~Kerley, K.~Kjems, E.~Knoepfel, V.~Koriakin, J.~Lo, C.~Lu,
  Z.~Mehring, A.~Moufarek, H.~Nandwani, V.~Oliveira, F.~Pardo, J.~Park,
  A.~Pierson, B.~Poole, H.~Ran, T.~Salimans, M.~Sanchez, I.~Saprykin, A.~Shen,
  S.~Sidhwani, D.~Smith, J.~Stanton, H.~Tomlinson, D.~Vijaykumar, L.~Wang,
  P.~Wingfield, N.~Wong, K.~Xu, C.~Yew, N.~Young, V.~Zubov, D.~Eck, D.~Erhan,
  K.~Kavukcuoglu, D.~Hassabis, Z.~Gharamani, R.~Hadsell, A.~van~den Oord,
  I.~Mosseri, A.~Bolton, S.~Singh, and T.~Rockt{\"a}schel.
\newblock Genie 3: A new frontier for world models.
\newblock 2025.
\newblock URL
  \url{https://deepmind.google/discover/blog/genie-3-a-new-frontier-for-world-models/}.

\bibitem[Berner et~al.(2019)Berner, Brockman, Chan, Cheung, D{\k{e}}biak,
  Dennison, Farhi, Fischer, Hashme, Hesse, et~al.]{berner2019dota}
C.~Berner, G.~Brockman, B.~Chan, V.~Cheung, P.~D{\k{e}}biak, C.~Dennison,
  D.~Farhi, Q.~Fischer, S.~Hashme, C.~Hesse, et~al.
\newblock Dota 2 with large scale deep reinforcement learning.
\newblock \emph{arXiv preprint arXiv:1912.06680}, 2019.

\bibitem[Bradbury et~al.(2018)Bradbury, Frostig, Hawkins, Johnson, Leary,
  Maclaurin, Necula, Paszke, Vander{P}las, Wanderman-{M}ilne, and
  Zhang]{jax2018github}
J.~Bradbury, R.~Frostig, P.~Hawkins, M.~J. Johnson, C.~Leary, D.~Maclaurin,
  G.~Necula, A.~Paszke, J.~Vander{P}las, S.~Wanderman-{M}ilne, and Q.~Zhang.
\newblock {JAX}: composable transformations of {P}ython+{N}um{P}y programs,
  2018.
\newblock URL \url{http://github.com/jax-ml/jax}.

\bibitem[Brown et~al.(2020)Brown, Mann, Ryder, Subbiah, Kaplan, Dhariwal,
  Neelakantan, Shyam, Sastry, Askell, et~al.]{brown2020language}
T.~Brown, B.~Mann, N.~Ryder, M.~Subbiah, J.~D. Kaplan, P.~Dhariwal,
  A.~Neelakantan, P.~Shyam, G.~Sastry, A.~Askell, et~al.
\newblock Language models are few-shot learners.
\newblock In \emph{Advances in neural information processing systems},
  volume~33, 2020.

\bibitem[Bruce et~al.(2024)Bruce, Dennis, Edwards, Parker-Holder, Shi, Hughes,
  Lai, Mavalankar, Steigerwald, Apps, Aytar, Bechtle, Behbahani, Chan, Heess,
  Gonzalez, Osindero, Ozair, Reed, Zhang, Zolna, Clune, de~Freitas, Singh, and
  Rockt{\"a}schel]{bruce2024genie}
J.~Bruce, M.~D. Dennis, A.~Edwards, J.~Parker-Holder, Y.~Shi, E.~Hughes,
  M.~Lai, A.~Mavalankar, R.~Steigerwald, C.~Apps, Y.~Aytar, S.~M.~E. Bechtle,
  F.~Behbahani, S.~C. Chan, N.~Heess, L.~Gonzalez, S.~Osindero, S.~Ozair,
  S.~Reed, J.~Zhang, K.~Zolna, J.~Clune, N.~de~Freitas, S.~Singh, and
  T.~Rockt{\"a}schel.
\newblock Genie: Generative interactive environments.
\newblock In \emph{Proceedings of the 41st International Conference on Machine
  Learning}, 2024.

\bibitem[Castricato et~al.(2025)Castricato, Matiana, Lapp, and
  BuGhanem]{wayfarer_labs_owl_wms}
L.~Castricato, S.~Matiana, A.~Lapp, and S.~BuGhanem.
\newblock {owl-wms: Basic world models}, 2025.
\newblock URL \url{https://github.com/Wayfarer-Labs/owl-wms}.

\bibitem[Chang et~al.(2022)Chang, Zhang, Jiang, Liu, and
  Freeman]{chang2022maskgit}
H.~Chang, H.~Zhang, L.~Jiang, C.~Liu, and W.~T. Freeman.
\newblock Maskgit: Masked generative image transformer.
\newblock In \emph{Proceedings of the IEEE/CVF conference on computer vision
  and pattern recognition}, 2022.

\bibitem[Chen et~al.(2024)Chen, Mart{\'\i}~Mons{\'o}, Du, Simchowitz, Tedrake,
  and Sitzmann]{chen2024diffusion}
B.~Chen, D.~Mart{\'\i}~Mons{\'o}, Y.~Du, M.~Simchowitz, R.~Tedrake, and
  V.~Sitzmann.
\newblock Diffusion forcing: Next-token prediction meets full-sequence
  diffusion.
\newblock In \emph{Advances in neural information processing systems},
  volume~37, 2024.

\bibitem[Chowdhery et~al.(2023)Chowdhery, Narang, Devlin, Bosma, Mishra,
  Roberts, Barham, Chung, Sutton, Gehrmann, et~al.]{chowdhery2022palm}
A.~Chowdhery, S.~Narang, J.~Devlin, M.~Bosma, G.~Mishra, A.~Roberts, P.~Barham,
  H.~W. Chung, C.~Sutton, S.~Gehrmann, et~al.
\newblock Palm: Scaling language modeling with pathways.
\newblock \emph{Journal of Machine Learning Research}, 24\penalty0 (240), 2023.

\bibitem[Christiano et~al.(2017)Christiano, Leike, Brown, Martic, Legg, and
  Amodei]{christiano2017deep}
P.~F. Christiano, J.~Leike, T.~Brown, M.~Martic, S.~Legg, and D.~Amodei.
\newblock Deep reinforcement learning from human preferences.
\newblock In \emph{Advances in neural information processing systems},
  volume~30, 2017.

\bibitem[Cobbe et~al.(2020)Cobbe, Hesse, Hilton, and
  Schulman]{cobbe2019leveraging}
K.~Cobbe, C.~Hesse, J.~Hilton, and J.~Schulman.
\newblock Leveraging procedural generation to benchmark reinforcement learning.
\newblock In \emph{Proceedings of the 37th International Conference on Machine
  Learning}, 2020.

\bibitem[{{Cursor}}(2025)]{cursor2025tab}
{{Cursor}}.
\newblock A new tab model, 2025.
\newblock URL \url{https://cursor.com/blog/tab-update}.

\bibitem[Dao et~al.(2022)Dao, Fu, Ermon, Rudra, and
  R{\'e}]{dao2022flashattention}
T.~Dao, D.~Fu, S.~Ermon, A.~Rudra, and C.~R{\'e}.
\newblock Flashattention: Fast and memory-efficient exact attention with
  io-awareness.
\newblock In \emph{Advances in neural information processing systems},
  volume~35, 2022.

\bibitem[Decart et~al.(2024)Decart, McIntyre, Campbell, Chen, and
  Wachen]{decart2024oasis}
E.~Decart, Q.~McIntyre, S.~Campbell, X.~Chen, and R.~Wachen.
\newblock Oasis: A universe in a transformer, 2024.
\newblock URL \url{https://oasis-model.github.io}.

\bibitem[Deng et~al.(2009)Deng, Dong, Socher, Li, Li, and
  Fei-Fei]{deng2009imagenet}
J.~Deng, W.~Dong, R.~Socher, L.-J. Li, K.~Li, and L.~Fei-Fei.
\newblock Imagenet: A large-scale hierarchical image database.
\newblock In \emph{Proceedings of the IEEE/CVF conference on computer vision
  and pattern recognition}, 2009.

\bibitem[Devlin et~al.(2019)Devlin, Chang, Lee, and Toutanova]{devlin2019bert}
J.~Devlin, M.-W. Chang, K.~Lee, and K.~Toutanova.
\newblock Bert: Pre-training of deep bidirectional transformers for language
  understanding.
\newblock In \emph{Proceedings of the 2019 conference of the North American
  chapter of the association for computational linguistics: human language
  technologies, volume 1 (long and short papers)}, 2019.

\bibitem[Fedus et~al.(2022)Fedus, Zoph, and Shazeer]{JMLR:v23:21-0998}
W.~Fedus, B.~Zoph, and N.~Shazeer.
\newblock Switch transformers: Scaling to trillion parameter models with simple
  and efficient sparsity.
\newblock \emph{Journal of Machine Learning Research}, 23\penalty0 (120), 2022.

\bibitem[Frans et~al.(2024)Frans, Hafner, Levine, and Abbeel]{frans2024one}
K.~Frans, D.~Hafner, S.~Levine, and P.~Abbeel.
\newblock One step diffusion via shortcut models.
\newblock \emph{arXiv preprint arXiv:2410.12557}, 2024.

\bibitem[Google(2024)]{ArrayRecord}
Google.
\newblock Arrayrecord: A file format for efficient io, 2024.
\newblock URL \url{https://github.com/google/array_record}.

\bibitem[Guo et~al.(2025{\natexlab{a}})Guo, Yang, Zhang, Song, Wang, Zhu, Xu,
  Zhang, Ma, Bi, et~al.]{guo2025deepseek}
D.~Guo, D.~Yang, H.~Zhang, J.~Song, P.~Wang, Q.~Zhu, R.~Xu, R.~Zhang, S.~Ma,
  X.~Bi, et~al.
\newblock Deepseek-r1 incentivizes reasoning in llms through reinforcement
  learning.
\newblock \emph{Nature}, 645\penalty0 (8081), 2025{\natexlab{a}}.

\bibitem[Guo et~al.(2025{\natexlab{b}})Guo, Ye, He, Wu, Jiang, Pearce, and
  Bian]{guo2025mineworld}
J.~Guo, Y.~Ye, T.~He, H.~Wu, Y.~Jiang, T.~Pearce, and J.~Bian.
\newblock Mineworld: a real-time and open-source interactive world model on
  minecraft.
\newblock \emph{arXiv preprint arXiv:2504.08388}, 2025{\natexlab{b}}.

\bibitem[Ha and Schmidhuber(2018)]{ha2018world}
D.~Ha and J.~Schmidhuber.
\newblock Recurrent world models facilitate policy evolution.
\newblock In \emph{Advances in neural information processing systems},
  volume~31, 2018.

\bibitem[Hafner et~al.(2019)Hafner, Lillicrap, Fischer, Villegas, Ha, Lee, and
  Davidson]{hafner2019learning}
D.~Hafner, T.~Lillicrap, I.~Fischer, R.~Villegas, D.~Ha, H.~Lee, and
  J.~Davidson.
\newblock Learning latent dynamics for planning from pixels.
\newblock In \emph{Proceedings of the 36th International Conference on Machine
  Learning}, 2019.

\bibitem[Hafner et~al.(2020{\natexlab{a}})Hafner, Lillicrap, Ba, and
  Norouzi]{hafner2019dream}
D.~Hafner, T.~Lillicrap, J.~Ba, and M.~Norouzi.
\newblock Dream to control: Learning behaviors by latent imagination.
\newblock In \emph{The Eighth International Conference on Learning
  Representations}, 2020{\natexlab{a}}.

\bibitem[Hafner et~al.(2020{\natexlab{b}})Hafner, Lillicrap, Norouzi, and
  Ba]{hafner2020mastering}
D.~Hafner, T.~Lillicrap, M.~Norouzi, and J.~Ba.
\newblock Mastering atari with discrete world models.
\newblock \emph{arXiv preprint arXiv:2010.02193}, 2020{\natexlab{b}}.

\bibitem[Hafner et~al.(2025{\natexlab{a}})Hafner, Pasukonis, Ba, and
  Lillicrap]{hafner2023mastering}
D.~Hafner, J.~Pasukonis, J.~Ba, and T.~Lillicrap.
\newblock Mastering diverse control tasks through world models.
\newblock \emph{Nature}, 2025{\natexlab{a}}.

\bibitem[Hafner et~al.(2025{\natexlab{b}})Hafner, Yan, and
  Lillicrap]{hafner2025training}
D.~Hafner, W.~Yan, and T.~Lillicrap.
\newblock Training agents inside of scalable world models.
\newblock \emph{arXiv preprint arXiv:2509.24527}, 2025{\natexlab{b}}.

\bibitem[He et~al.(2022)He, Chen, Xie, Li, Doll{\'a}r, and
  Girshick]{he2022masked}
K.~He, X.~Chen, S.~Xie, Y.~Li, P.~Doll{\'a}r, and R.~Girshick.
\newblock Masked autoencoders are scalable vision learners.
\newblock In \emph{Proceedings of the IEEE/CVF conference on computer vision
  and pattern recognition}, 2022.

\bibitem[Ho et~al.(2019)Ho, Kalchbrenner, Weissenborn, and
  Salimans]{ho2019axial}
J.~Ho, N.~Kalchbrenner, D.~Weissenborn, and T.~Salimans.
\newblock Axial attention in multidimensional transformers.
\newblock \emph{arXiv preprint arXiv:1912.12180}, 2019.

\bibitem[Hu et~al.(2023)Hu, Russell, Yeo, Murez, Fedoseev, Kendall, Shotton,
  and Corrado]{hu2023gaia}
A.~Hu, L.~Russell, H.~Yeo, Z.~Murez, G.~Fedoseev, A.~Kendall, J.~Shotton, and
  G.~Corrado.
\newblock Gaia-1: A generative world model for autonomous driving.
\newblock \emph{arXiv preprint arXiv:2309.17080}, 2023.

\bibitem[Johnson(2024)]{johnson2024penzai}
D.~D. Johnson.
\newblock {Penzai} + {Treescope}: A toolkit for interpreting, visualizing, and
  editing models as data.
\newblock \emph{ICML 2024 Workshop on Mechanistic Interpretability}, 2024.

\bibitem[Jozefowicz et~al.(2016)Jozefowicz, Vinyals, Schuster, Shazeer, and
  Wu]{jozefowicz2016exploring}
R.~Jozefowicz, O.~Vinyals, M.~Schuster, N.~Shazeer, and Y.~Wu.
\newblock Exploring the limits of language modeling.
\newblock \emph{arXiv preprint arXiv:1602.02410}, 2016.

\bibitem[Li et~al.(2025)Li, Tang, Xu, Wu, Zhou, Shao, Yu, Cao, and
  Lu]{li2025hunyuan}
J.~Li, J.~Tang, Z.~Xu, L.~Wu, Y.~Zhou, S.~Shao, T.~Yu, Z.~Cao, and Q.~Lu.
\newblock Hunyuan-gamecraft: High-dynamic interactive game video generation
  with hybrid history condition.
\newblock \emph{arXiv preprint arXiv:2506.17201}, 2025.

\bibitem[Lin and Cheng(2025)]{LinCheng2025GeminiICPC}
H.~M. Lin and H.-T. Cheng.
\newblock Gemini achieves gold-level performance at the international
  collegiate programming contest world finals, 2025.
\newblock URL
  \url{https://deepmind.google/discover/blog/gemini-achieves-gold-level-performance-at-the-international-collegiate-programming-contest-world-finals/}.

\bibitem[Luong and Lockhart(2025)]{LuongLockhart2025GeminiIMO}
T.~Luong and E.~Lockhart.
\newblock Advanced version of gemini with deep think officially achieves
  gold-medal standard at the international mathematical olympiad, 2025.
\newblock URL
  \url{https://deepmind.google/discover/blog/advanced-version-of-gemini-with-deep-think-officially-achieves-gold-medal-standard-at-the-international-mathematical-olympiad/}.

\bibitem[Mahajan et~al.(2018)Mahajan, Girshick, Ramanathan, He, Paluri, Li,
  Bharambe, and Van Der~Maaten]{mahajan2018exploring}
D.~Mahajan, R.~Girshick, V.~Ramanathan, K.~He, M.~Paluri, Y.~Li, A.~Bharambe,
  and L.~Van Der~Maaten.
\newblock Exploring the limits of weakly supervised pretraining.
\newblock In \emph{Proceedings of the European conference on computer vision},
  2018.

\bibitem[{NVIDIA Corporation}(2025)]{NVIDIA_cuDNN_Attention}
{NVIDIA Corporation}.
\newblock cudnn frontend api: Scaled dot product attention fp16/bf16 forward,
  2025.
\newblock URL
  \url{https://docs.nvidia.com/deeplearning/cudnn/frontend/latest/operations/Attention.html}.

\bibitem[{OpenXLA Project}(2025)]{openxla-shardy}
{OpenXLA Project}.
\newblock Shardy: an mlir-based tensor partitioning system, 2025.
\newblock URL \url{https://github.com/openxla/shardy}.

\bibitem[Parker-Holder et~al.(2022)Parker-Holder, Jiang, Dennis, Samvelyan,
  Foerster, Grefenstette, and Rockt{\"a}schel]{parker2022evolving}
J.~Parker-Holder, M.~Jiang, M.~Dennis, M.~Samvelyan, J.~Foerster,
  E.~Grefenstette, and T.~Rockt{\"a}schel.
\newblock Evolving curricula with regret-based environment design.
\newblock In \emph{Proceedings of the 39th International Conference on Machine
  Learning}, 2022.

\bibitem[Parker-Holder et~al.(2024)Parker-Holder, Ball, Bruce, Dasagi,
  Holsheimer, Kaplanis, Moufarek, Scully, Shar, Shi, Spencer, Yung, Dennis,
  Kenjeyev, Long, Mnih, Chan, Gazeau, Li, Pardo, Wang, Zhang, Besse, Harley,
  Mitenkova, Wang, Clune, Hassabis, Hadsell, Bolton, Singh, and
  Rockt{\"a}schel]{parkerholder2024genie2}
J.~Parker-Holder, P.~Ball, J.~Bruce, V.~Dasagi, K.~Holsheimer, C.~Kaplanis,
  A.~Moufarek, G.~Scully, J.~Shar, J.~Shi, S.~Spencer, J.~Yung, M.~Dennis,
  S.~Kenjeyev, S.~Long, V.~Mnih, H.~Chan, M.~Gazeau, B.~Li, F.~Pardo, L.~Wang,
  L.~Zhang, F.~Besse, T.~Harley, A.~Mitenkova, J.~Wang, J.~Clune, D.~Hassabis,
  R.~Hadsell, A.~Bolton, S.~Singh, and T.~Rockt{\"a}schel.
\newblock Genie 2: A large-scale foundation world model.
\newblock 2024.
\newblock URL
  \url{https://deepmind.google/discover/blog/genie-2-a-large-scale-foundation-world-model/}.

\bibitem[Pearce et~al.(2025)Pearce, Rashid, Bignell, Georgescu, Devlin, and
  Hofmann]{pearce2024scaling}
T.~Pearce, T.~Rashid, D.~Bignell, R.~Georgescu, S.~Devlin, and K.~Hofmann.
\newblock Scaling laws for pre-training agents and world models.
\newblock In \emph{Proceedings of the 42nd International Conference on Machine
  Learning}, 2025.

\bibitem[Peebles and Xie(2023)]{peebles2023scalable}
W.~Peebles and S.~Xie.
\newblock Scalable diffusion models with transformers.
\newblock In \emph{Proceedings of the IEEE/CVF conference on computer vision
  and pattern recognition}, 2023.

\bibitem[Radford et~al.(2018)Radford, Narasimhan, Salimans, Sutskever,
  et~al.]{radford2018improving}
A.~Radford, K.~Narasimhan, T.~Salimans, I.~Sutskever, et~al.
\newblock Improving language understanding by generative pre-training.
\newblock \emph{OpenAI blog}, 2018.

\bibitem[Radford et~al.(2019)Radford, Wu, Child, Luan, Amodei, Sutskever,
  et~al.]{radford2019language}
A.~Radford, J.~Wu, R.~Child, D.~Luan, D.~Amodei, I.~Sutskever, et~al.
\newblock Language models are unsupervised multitask learners.
\newblock \emph{OpenAI blog}, 2019.

\bibitem[Radford et~al.(2021)Radford, Kim, Hallacy, Ramesh, Goh, Agarwal,
  Sastry, Askell, Mishkin, Clark, et~al.]{radford2021learning}
A.~Radford, J.~W. Kim, C.~Hallacy, A.~Ramesh, G.~Goh, S.~Agarwal, G.~Sastry,
  A.~Askell, P.~Mishkin, J.~Clark, et~al.
\newblock Learning transferable visual models from natural language
  supervision.
\newblock In \emph{Proceedings of the 38th International Conference on Machine
  Learning}, 2021.

\bibitem[Raffel et~al.(2020)Raffel, Shazeer, Roberts, Lee, Narang, Matena,
  Zhou, Li, and Liu]{raffel2020exploring}
C.~Raffel, N.~Shazeer, A.~Roberts, K.~Lee, S.~Narang, M.~Matena, Y.~Zhou,
  W.~Li, and P.~J. Liu.
\newblock Exploring the limits of transfer learning with a unified text-to-text
  transformer.
\newblock \emph{Journal of Machine Learning Research}, 21\penalty0 (140), 2020.

\bibitem[Ritter et~al.(2023)Ritter, Indyk, Singh, Audibert, Seelam, Hanes, Lau,
  Olesiak, Kang, and Wu]{grain2023github}
M.~Ritter, I.~Indyk, A.~Singh, A.~Audibert, A.~Seelam, C.~Hanes, E.~Lau,
  J.~Olesiak, J.~Kang, and X.~Wu.
\newblock {Grain} - feeding jax models, 2023.
\newblock URL \url{http://github.com/google/grain}.

\bibitem[Salmon et~al.(2011)Salmon, Moraes, Dror, and Shaw]{salmon2011parallel}
J.~K. Salmon, M.~A. Moraes, R.~O. Dror, and D.~E. Shaw.
\newblock Parallel random numbers: as easy as 1, 2, 3.
\newblock In \emph{Proceedings of 2011 international conference for high
  performance computing, networking, storage and analysis}, 2011.

\bibitem[Savov et~al.(2025)Savov, Kazemi, Mahdi, Paudel, Wang, and
  Van~Gool]{Savov_2025_CVPR}
N.~Savov, N.~Kazemi, M.~Mahdi, D.~P. Paudel, X.~Wang, and L.~Van~Gool.
\newblock Exploration-driven generative interactive environments.
\newblock In \emph{Proceedings of the IEEE/CVF conference on computer vision
  and pattern recognition}, 2025.

\bibitem[Schmidt and Jiang(2024)]{schmidt2023learning}
D.~Schmidt and M.~Jiang.
\newblock Learning to act without actions.
\newblock In \emph{The Twelfth International Conference on Learning
  Representations}, 2024.

\bibitem[Seid and Hojel(2024)]{lucid2024lucidv1}
R.~Seid and A.~Hojel.
\newblock Lucid v1: Real-time latent world models, 2024.
\newblock URL \url{https://www.lucid.ai/notes/lucid-v1}.

\bibitem[Shazeer(2024)]{shazeer2024shape}
N.~Shazeer.
\newblock Shape suffixes, 2024.
\newblock URL
  \url{https://medium.com/@NoamShazeer/shape-suffixes-good-coding-style-f836e72e24fd}.

\bibitem[Shoeybi et~al.(2019)Shoeybi, Patwary, Puri, LeGresley, Casper, and
  Catanzaro]{megatron-lm}
M.~Shoeybi, M.~Patwary, R.~Puri, P.~LeGresley, J.~Casper, and B.~Catanzaro.
\newblock Megatron-lm: Training multi-billion parameter language models using
  model parallelism.
\newblock \emph{arXiv preprint arXiv:1909.08053}, 2019.

\bibitem[Silver et~al.(2016)Silver, Huang, Maddison, Guez, Sifre, Van
  Den~Driessche, Schrittwieser, Antonoglou, Panneershelvam, Lanctot,
  et~al.]{silver2016mastering}
D.~Silver, A.~Huang, C.~J. Maddison, A.~Guez, L.~Sifre, G.~Van Den~Driessche,
  J.~Schrittwieser, I.~Antonoglou, V.~Panneershelvam, M.~Lanctot, et~al.
\newblock Mastering the game of go with deep neural networks and tree search.
\newblock \emph{Nature}, 529\penalty0 (7587), 2016.

\bibitem[Srambical(2024)]{srambical2024going}
F.~Srambical.
\newblock Going beyond the causal mask in language modeling.
\newblock \emph{p(doom) blog}, 2024.
\newblock URL \url{https://pdoom.org/blog.html}.

\bibitem[Srambical and Mahajan(2025)]{srambical2025crowd-sourcing}
F.~Srambical and M.~Mahajan.
\newblock Crowd-sourcing a dataset to make agents code like humans.
\newblock \emph{p(doom) blog}, 2025.
\newblock URL \url{https://pdoom.org/blog.html}.

\bibitem[Sutton(1991)]{sutton1991dyna}
R.~S. Sutton.
\newblock Dyna, an integrated architecture for learning, planning, and
  reacting.
\newblock \emph{ACM Sigart Bulletin}, 2\penalty0 (4), 1991.

\bibitem[Valevski et~al.(2025)Valevski, Leviathan, Arar, and
  Fruchter]{valevski2024diffusion}
D.~Valevski, Y.~Leviathan, M.~Arar, and S.~Fruchter.
\newblock Diffusion models are real-time game engines.
\newblock In \emph{The Thirteenth International Conference on Learning
  Representations}, 2025.

\bibitem[Van Den~Oord et~al.(2017)Van Den~Oord, Vinyals, and
  Kavukcuoglu]{van2017neural}
A.~Van Den~Oord, O.~Vinyals, and K.~Kavukcuoglu.
\newblock Neural discrete representation learning.
\newblock In \emph{Advances in neural information processing systems},
  volume~30, 2017.

\bibitem[Willi et~al.(2024)Willi, Jackson, and Foerster]{willi2024jafar}
T.~Willi, M.~T. Jackson, and J.~N. Foerster.
\newblock Jafar: An open-source genie reimplementation in jax.
\newblock In \emph{First Workshop on Controllable Video Generation @ ICML
  2024}, 2024.

\bibitem[Witten and {{MaxText Authors}}(2024)]{maxtext-library}
R.~Witten and {{MaxText Authors}}.
\newblock {MaxText: A simple, performant and scalable Jax LLM}, 2024.
\newblock URL \url{https://github.com/google/maxtext}.

\bibitem[Zhai et~al.(2022)Zhai, Kolesnikov, Houlsby, and
  Beyer]{zhai2022scaling}
X.~Zhai, A.~Kolesnikov, N.~Houlsby, and L.~Beyer.
\newblock Scaling vision transformers.
\newblock In \emph{Proceedings of the IEEE/CVF conference on computer vision
  and pattern recognition}, 2022.

\bibitem[Zhang et~al.(2018)Zhang, Isola, Efros, Shechtman, and
  Wang]{zhang2018unreasonable}
R.~Zhang, P.~Isola, A.~A. Efros, E.~Shechtman, and O.~Wang.
\newblock The unreasonable effectiveness of deep features as a perceptual
  metric.
\newblock In \emph{Proceedings of the IEEE/CVF conference on computer vision
  and pattern recognition}, 2018.

\end{thebibliography}
\appendix

\begin{figure}
  \centering
  \includegraphics[width=\textwidth]{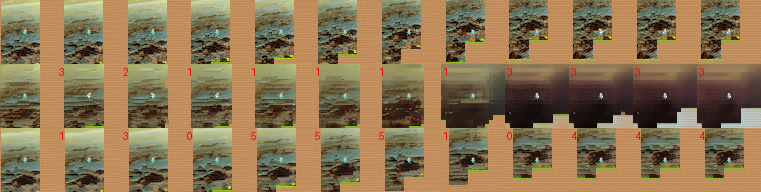}
  \caption{Autoregressive sampling of Jasmine when adding (middle row) and prepending actions (bottom row) on the CoinRun case study with four conditioning frames (conditioning frames not shown). 
  The top row shows the ground-truth sequence. 
}
\label{fig:sampling_prepend_vs_no_prepend}
\end{figure}

\section{Coinrun Case Study}
\label{sec:coinrun-case-study}
For the CoinRun case study, we strictly adhere to the setting of \citet{bruce2024genie} and train our models to unmask sequences of 16 frames with a resolution of 64x64 pixels per frame. To generate the dataset, we capture 50M observation frames and corresponding ground-truth actions during random agent rollouts and only use the ground-truth actions for a LAM ablation (\Cref{sec:ablation-gt}). Instead of sampling seeds from a fixed pool as described in \citet{bruce2024genie}, we initialize all episodes with a seed unique to the respective episode. Furthermore, we verify that our generated dataset contains no duplicate episodes and only find 7.46\% duplicate frames. 
We confirm that the validation and test set are disjoint from the train set and publish our script for duplication detection along with the repository\footnote{\url{{https://github.com/p-doom/jasmine/blob/main/data/jasmine_data/detect_array_record_duplicates.py}}}.
While train metrics are near-identical between Genie's configuration and our action-prepending modification, rollout quality differs significantly (\Cref{fig:case-study-plots}). We collect rollout metrics during training (\Cref{sec:experiment-metrics}) that capture this discrepancy.

\section{Ablations}
For our ablations, we reuse the CoinRun setting and ablate from Jasmine's base configuration, depicted in \Cref{tab:hparams}.
\label{sec:arch-ablations}
\begin{figure}
  \centering
  \includegraphics[width=1.0\textwidth]{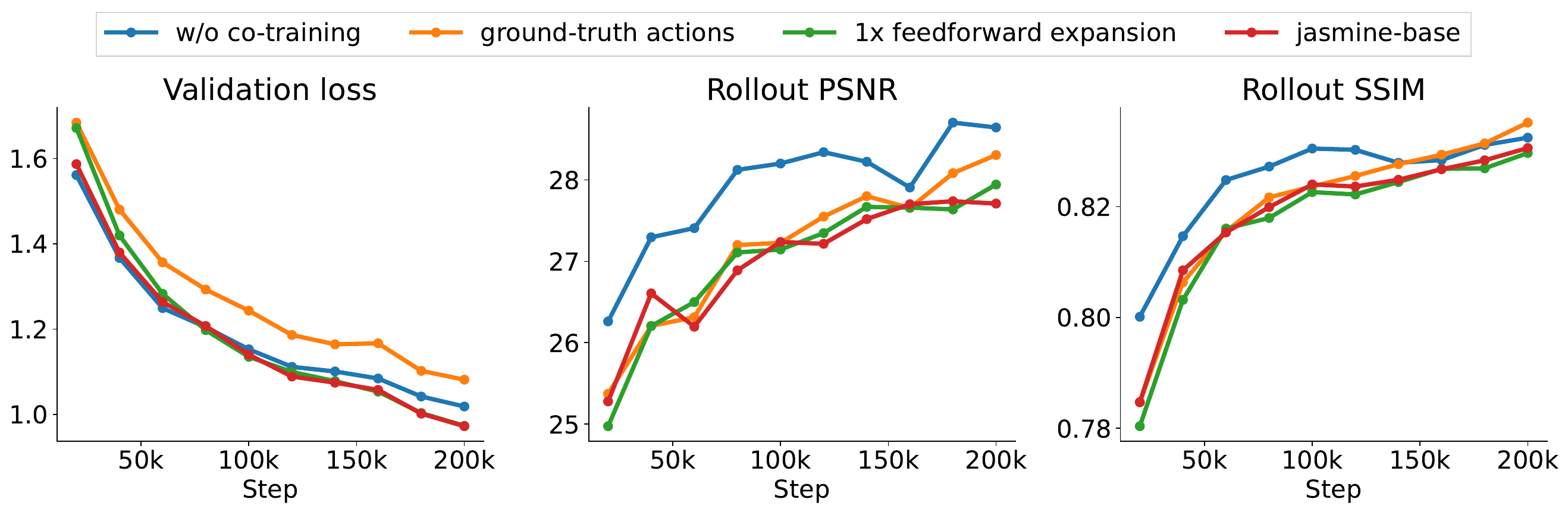}
  \caption{Architectural ablations of Jasmine's base configuration (refer to \Cref{tab:hparams}) on CoinRun. We report loss (left) and rollout metrics (middle and right) of the dynamics model on a validation set.}
  \label{fig:arch-ablations}
\end{figure}

\paragraph{Co-training LAM and dynamics model}
\label{sec:ablation-co-training}
\citet{bruce2024genie} co-train the LAM and the dynamics model. However, their implementation remains unclear as the LAM is supervised on frames while the dynamics model is supervised on tokens. One approach to co-training is a combined loss function including stop-gradients that prevent gradients from flowing from the dynamics model to the LAM. However, such a combined loss formulation remains unmentioned in \citet{bruce2024genie}. 
\citet{willi2024jafar} instead train LAM and dynamics model sequentially, thus reducing memory footprint at the cost of longer total training time. 
For Jasmine's co-training implementation, we omit the LAM decoder entirely and allow gradients to flow from the dynamics model to the LAM.

\paragraph{Training with ground-truth actions}
\label{sec:ablation-gt}
We ablate the LAM (Figure \ref{fig:arch-ablations}) by training the dynamics model using ground-truth actions captured in the environment. We use an embedding layer to map action indices to action latents, which are then used as additional input to the dynamics model.

\begin{figure}
  \centering
  \includegraphics[width=1.0\textwidth]{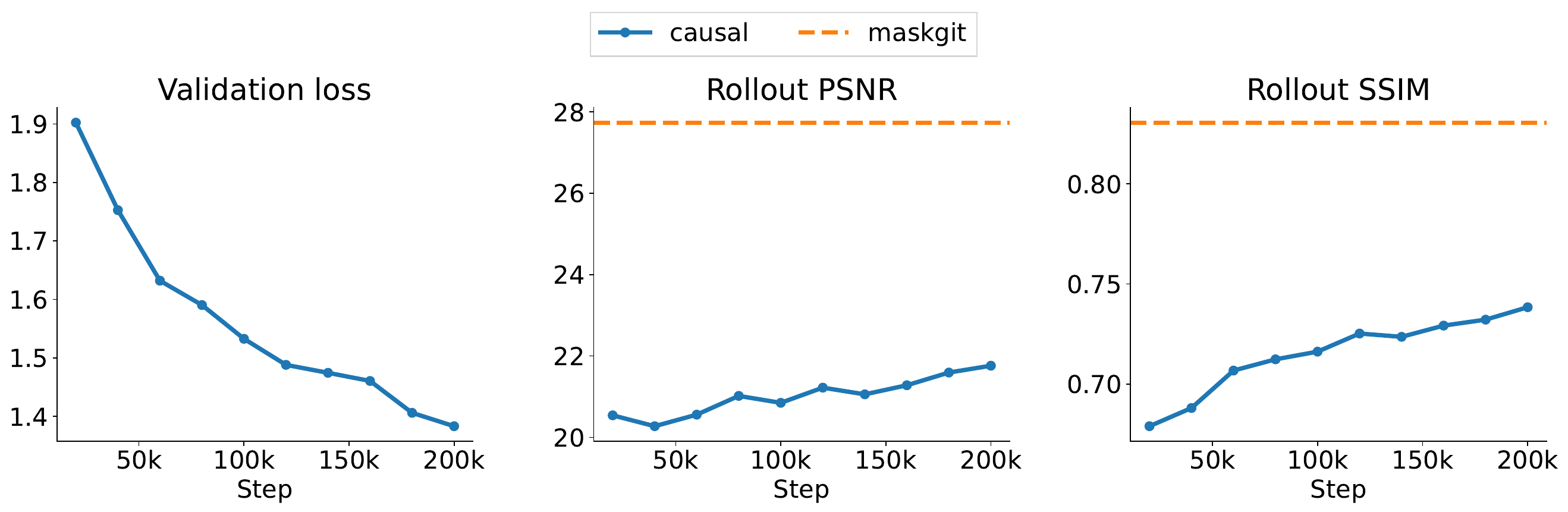}
  \caption{Loss (left) and rollout metrics (middle and right) of the fully causal baseline. We depict the final performance of our MaskGIT implementation (Jasmine's default configuration) for the rollout metrics. The loss indicates that the causal baseline might benefit from longer training and separately tuned hyperparameters. The losses between the two architectures are not comparable, hence we omit the MaskGIT loss.}
  \label{fig:causal-ablations}
\end{figure}

\paragraph{Throughput ablations}
\label{sec:throughput-ablation}

We ablate core components of Jasmine's infrastructure in Table \ref{tab:infra-ablations}. Replacing Grain with the default data loader of \citet{willi2024jafar} reduces throughput by an order of magnitude. In Jasmine's base CoinRun configuration (sub-100M parameter model and a maximum attention sequence length of 16) the XLA compiler dispatches higher-throughput CUDA kernels than FlashAttention. While FlashAttention only outperforms XLA-compiled kernels at large model sizes and sequence lengths (\Cref{tab:big-model-flash-attn-ablation}), we enable it by default to reduce accelerator memory usage. At small batch sizes, the compiled train loop of our configuration achieves higher throughput using full precision. We attribute this to XLA largely operating based on heuristics, and posit that writing optimized Pallas kernels for key operations in the model forward pass will result in mixed precision outperforming full precision in throughput, even at small batch sizes. 

The ArrayRecord file format allows storing a configurable amount of records per file. In our case, each record corresponds to a sequence of frames and actions. We find that the chosen format significantly affects throughput (Tables \ref{tab:chunking-formats-1} and \ref{tab:chunking-formats-2}). Based on preliminary experiments, we preprocess the dataset to have 100 records per ArrayRecord file with 160 frames per record.

\begin{table}
\begin{center}
\begin{tabular}{lcc}
\toprule
 & \textbf{Throughput (bs=36)} & \textbf{Throughput (bs=2048)} \\
\midrule
Jasmine-base & 1.00x	            & 1.00x \\
w/o grain data loader & 0.25x     & 0.11x \\
w/o flash attention    & 1.15x	& 1.04x \\
w/o mixed precision & 1.18x	& 0.71x \\
\bottomrule
\end{tabular}
\caption{Training throughput of infrastructure ablations, relative to Jasmine-base. We report the throughput at Genie's default batch size (36), as well as at the batch size resulting in the highest throughput (2048) on a single H100 with 80GB of accelerator memory.}
\label{tab:infra-ablations}
\end{center}
\end{table}

\begin{table}
\begin{center}
\begin{tabular}{lcc}
\toprule
 & \textbf{Throughput (frames/sec)} & \textbf{Throughput (relative)} \\
\midrule
w/ flash attention & 36.15	         &  1.00x \\
w/o flash attention & 24.24 &  0.67x \\
\bottomrule
\end{tabular}
\caption{Training throughput using a larger model (1B parameter) and spatial sequence length (1024). We decrease the patch size to two. In this regime, FlashAttention yields higher throughput than the XLA compiler.}
\label{tab:big-model-flash-attn-ablation}
\end{center}
\end{table}

\begin{table}
\begin{center}
\begin{tabular}{llcc}
\toprule
\textbf{\# frames per record} & \textbf{\# records per file} & \textbf{Throughput (frames/sec)} & \textbf{Throughput (relative)}   \\
\midrule
16          & 100 & 7,527.27 & 1.12x \\
160 (Base)  & 100 & 6,709.09 & 1.00x \\
1,600       & 100 & 3,752.73 & 0.56x \\
16,000      & 100 & 3,720.00 & 0.55x \\
160,000     & 100 & 3,785.45 & 0.56x \\

\bottomrule
\end{tabular}
\caption{Training throughput at Genie's default batch size (36) with different number of frames per record. Throughput decreases as the number of frames per record increases. We opt for 160 frames per record to be able to vary the sequence length.} 
\label{tab:chunking-formats-1}
\end{center}
\end{table}

\begin{table}
\begin{center}
\begin{tabular}{llcc}
\toprule
\textbf{\# frames per record} & \textbf{\# records per file} & \textbf{Throughput (frames/sec)} & \textbf{Throughput (relative)}\\
\midrule
160         & 1         & 6,098.18 & 0.91x \\
160         & 10        & 6,643.64 & 0.99x \\
160 (Base)  & 100       & 6,709.09 & 1.00x \\
160         & 1,000     & 6,480.00 & 0.97x \\
160         & 10,000    & 6,763.64 & 1.01x \\
\bottomrule
\end{tabular}
\caption{Training throughput at Genie's default batch size (36) with different number of records per file.} 
\label{tab:chunking-formats-2}
\end{center}
\end{table}

\section{Diffusion Baseline}
\label{sec:diffusion-baseline}
We implement a diffusion baseline inspired by the Dreamer 4 architecture \citep{hafner2025training}, combining a masked autoencoder (MAE, \citet{he2022masked}) tokenizer  with an ST-DiT \citep{ho2019axial,peebles2023scalable} dynamics model trained under the diffusion-forcing objective \citep{chen2024diffusion}.
We leave implementing the shortcut objective \citep{frans2024one} to future work. 
\paragraph{Tokenizer} Following \citet{hafner2025training}, we use a MAE to compress raw video frames into continuous latents. Our autoencoder implementation uses an ST-Transformer backbone and a latent bottleneck. Before passing the latents to the decoder, we apply the tanh activation to constrain them to the range $(-1, 1)$ for downstream dynamics model training. We uniformly sample per-frame masking probabilities $p_i \sim U(0, 0.9)$. Unlike \citet{hafner2025training}, we omit the auxiliary LPIPS loss \citep{zhang2018unreasonable} and directly train on pixel-level reconstructions using mean-squared error. We find the tokenizer hyperparameters from \Cref{tab:hparams} to work well for MAE training as well.

\paragraph{Dynamics model}
We implement diffusion forcing \citep{chen2024diffusion}, sampling an independent noise level per frame during training. Analogous to \citet{hafner2025training} and Jasmine-base, we prepend latent actions and the embedded denoising step to the patch latents. Following \citet{hafner2025training}, we use x-prediction\footnote{In the diffusion literature, x-prediction often refers to supervision in latent-space rather than pixel-space. We follow that nomenclature but believe that the term z-prediction is a more accurate description.} and employ a ramp loss. During inference, frame-wise latents are autoregressively generated with 25 denoising steps per frame, while past input latents are slightly corrupted using a noise level of 0.1. We adopt the hyperparameters of Jasmine-base, and only change the learning rate to 1e-4 following \citet{peebles2023scalable}.

\begin{figure}
  \centering
  \includegraphics[width=1.0\textwidth]{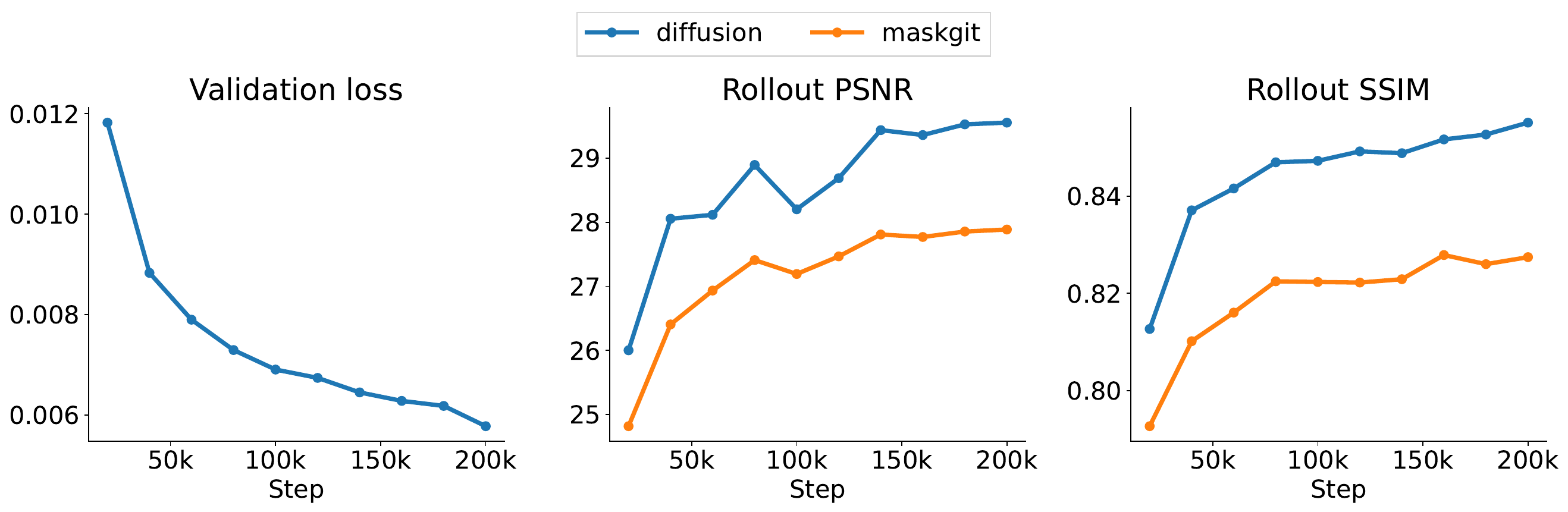}
  \caption{Loss (left) and rollout metrics (middle and right) of the diffusion baseline. We omit the MaskGIT loss as the losses are not comparable between the two architectures.}
  \label{fig:diffusion-plot}
\end{figure}

\section{Bitwise Determinism}
\label{sec:bitwise_deterministic}
On TPUs, bitwise determinism is guaranteed by Jasmine via proper usage of JAX's implementation of parallel random number generation via Threefry counters \citep{salmon2011parallel}. On GPUs however, an additional XLA flag (\texttt{xla\_gpu\_deterministic\_ops=true}) is needed in certain cases to guarantee identical training curves.

\section{Hyperparameter Configurations}
\label{sec:default-config}
We mention four distinct training configurations. We present hyperparameters and training settings of each configuration in \Cref{tab:hparams} and briefly describe them:

\begin{itemize}
    \item \textbf{Genie}: The hyperparameter configuration of Appendix F of \citet{bruce2024genie}, but with a patch size of 16 (for easier comparison against Jafar). We use this configuration for the CoinRun case study (\Cref{fig:sampling_prepend_vs_no_prepend}, middle row; refer to \Cref{sec:coinrun-case-study}).
    \item \textbf{Genie w/ prepend}: As detailed in \Cref{sec:jasmine}, we found a minimal modification to the Genie configuration to be necessary to yield generations faithful to the CoinRun environment (Figure \ref{fig:sampling}, bottom row and \Cref{fig:sampling_prepend_vs_no_prepend}, bottom row). This setting is identical to our Genie configuration, with the exception that we prepend latent actions to the video embeddings instead of adding them.
    \item \textbf{Jafar}: The hyperparameters used by Jafar \citep{willi2024jafar} for their CoinRun case study. This setting is identical to \citet{bruce2024genie}, but \citet{willi2024jafar} use a patch size of 16 for faster training. Unlike \citet{bruce2024genie}, they pre-train the LAM as opposed to co-training the LAM with the dynamics model. We use this configuration for the Jafar baseline runs (Figure \ref{fig:sampling}, middle row and \Cref{fig:wc_time}). We solely run this configuration with the Jafar repository.
    \item \textbf{Jasmine-base}: We define a base configuration for Jasmine that represents a trade-off between training speed, modeling quality and simplicity, integrating best practices from the language modeling literature. This is Jasmine's configuration in the wall-clock convergence comparison (\Cref{fig:wc_time}) between Jasmine and Jafar, as well as our architectural and infrastructure ablations (Figures \ref{fig:arch-ablations}, \ref{fig:sampling_multi} and Tables \ref{tab:ffn-ablation}, \ref{tab:infra-ablations}, \ref{tab:big-model-flash-attn-ablation}, \ref{tab:chunking-formats-1}, \ref{tab:chunking-formats-2}).
\end{itemize}

\begin{table}
\begin{center}
\begin{tabular}{ll| r|r|r|r}
\toprule
 & \textbf{Parameter} & \textbf{Genie} & \textbf{Genie w/ prepend} & \textbf{Jafar} & \textbf{Jasmine-base} \\ 
\midrule
\midrule
Tokenizer & \# blocks    & 8      & & & 4 \\
          & \# heads     & 8      & & & \\
          & model dim       & 512    & & & \\
          & ffn dim       & 512 & & & 2048 \\
          & \# codes & 1024 & & & \\
          & latent dim & 32 & & & \\
          & patch size & 16 & & & \\
          & total train steps  & 300k  & & & \\
          & learning rate   & $3 * 10^{-4}$  & & & \\
          & lr decay end & $3 * 10^{-4}$ & & &  0 \\ 
          & batch size & 48 & & & \\
\midrule
LAM       & \# blocks    & 8      & & & 4 \\
          & \# heads     & 8      & & & \\
          & model dim       & 512    & & & \\
          & ffn dim       & 512 & & & 2048 \\
          & \# codes & 6 & & & \\
          & latent dim & 32 & & & \\
          & patch size & 16 & & & \\
          & total train steps  & 200k  & & & \\
          & learning rate   & $3 * 10^{-5}$ & & & \\
          & lr decay end & $3 * 10^{-6}$ & & & 0 \\ 
          & batch size   & 48 & & & \\
\midrule
Dynamics  & \# blocks    & 12      & & & 6 \\
          & \# heads     & 8      & & & \\
          & model dim       & 512    & & & \\
          & ffn dim       & 512 & & & 2048 \\
          & total train steps  & 200k  & & & \\
          & learning rate   & $3 * 10^{-5}$  & & & \\
          & lr decay end & $3 * 10^{-6}$ & & & 0 \\ 
          & batch size   & 36 & & & \\
          & action conditioning & additive & prepend & & prepend \\
          & baseline & MaskGIT & & & \\
\midrule
\midrule
Training & optimizer    & AdamW      & & & \\
          & lr schedule & cos     & & & wsd \\
          & warmup steps  & 1k & & & \\
          & wsd decay steps & - & & & 10\% \\
          & dataset size (frames) & 50M & & & \\
          & co-training & yes & & no & \\
\midrule
\midrule
Inference & temperature & 1.0 & & & \\
          & maskgit steps & 25 & & & \\
\bottomrule
\end{tabular}
  \caption{Configurations used in our experiments. We only show the difference to our base Genie configuration.
  Note that \citet{bruce2024genie} uses a tokenizer patch size of four, and that we use an expanded dataset with 50M frames for all of our runs (to ensure that no method performs worse due to overfitting).}
  \label{tab:hparams}
\end{center}
\end{table}

\section{Extending MaskGIT to Videos}
\label{sec:maskgit_to_videos}
MaskGIT \citep{chang2022maskgit} is defined on images and there are multiple ways to extend it to videos. We follow \citet{willi2024jafar} by randomly masking tokens in the entire sequence using the uniformly sampled probability $p \sim U(0.5,1)$. An alternative would be to sample a different masking probability per frame, similar to \citet{chen2024diffusion}, or leaving $k \sim U(0,T)$ frames unmasked to closely emulate inference.

\section{Evaluation Metrics}
\label{sec:experiment-metrics}
\begin{figure}
  \centering
  \includegraphics[width=\textwidth, trim=1 0 0 0, clip]{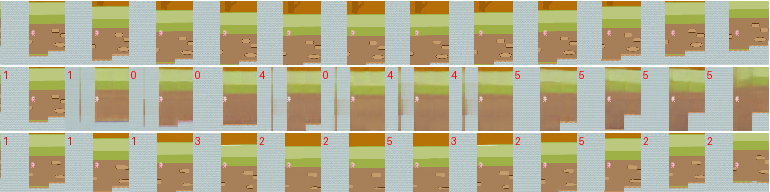}
  \includegraphics[width=\textwidth, trim=1 0 0 0, clip]{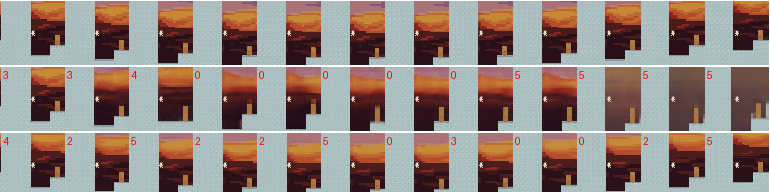}
  \includegraphics[width=\textwidth, trim=1 0 0 0, clip]{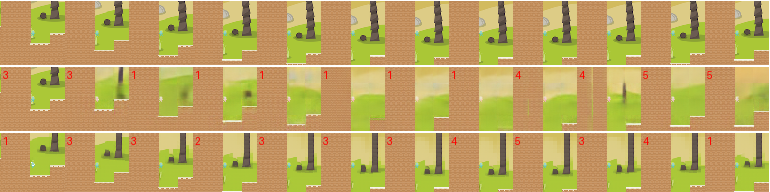}
  \includegraphics[width=\textwidth, trim=1 0 0 0, clip]{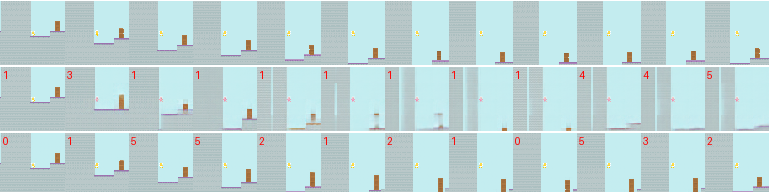}
  \includegraphics[width=\textwidth, trim=1 0 0 0, clip]{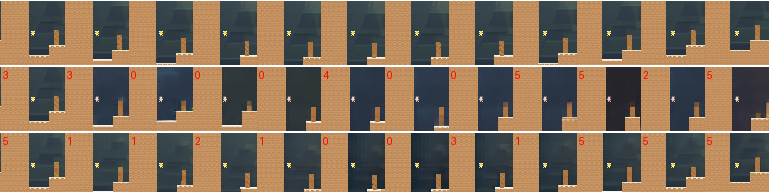}
  \caption{Autoregressive rollouts on five randomly selected trajectories from the CoinRun environment. Each set of three rows corresponds to one trajectory, showing ground-truth frames (top), Jafar samples (middle), and Jasmine samples (bottom). The four conditioning frames are omitted.}
  \label{fig:sampling_multi}
\end{figure}

\begin{figure}
  \centering
  \includegraphics[width=\textwidth, trim=1 0 0 0, clip]{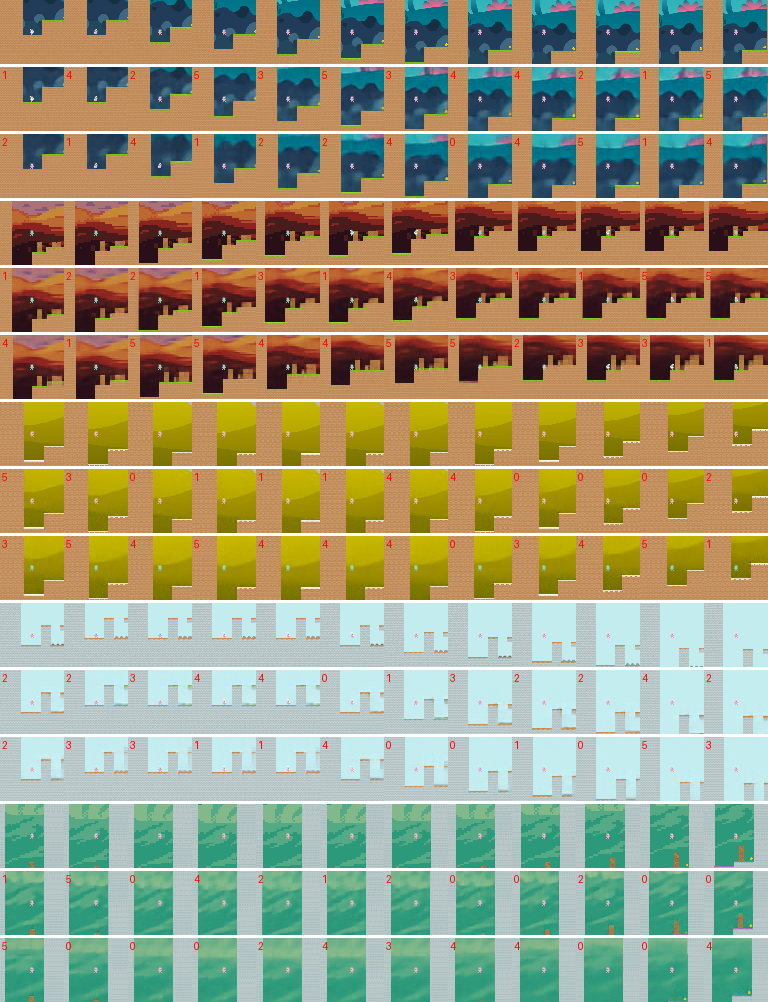}
  \caption{Autoregressive rollouts on five randomly selected trajectories from the CoinRun environment. Each set of three rows corresponds to one trajectory, showing ground-truth frames (top), samples using the diffusion baseline (middle), and samples using the MaskGIT baseline (bottom). The four conditioning frames are omitted.}
  \label{fig:sampling_diffusion}
\end{figure}

In early experiments, we found the Genie configuration to suffer from a discrepancy in performance between validation loss and autoregressive rollouts (Figure \ref{fig:case-study-plots}). Therefore, besides validation loss, Jasmine also tracks rollout metrics throughout training of the dynamics model. We generate a single frame using the sampling logic of the respective architecture and calculate SSIM and PSNR between the generated frame and the ground-truth. 
Although rollout metrics are calculated on a single frame, we find that they directly correlate with the model's performance in generating full rollouts. 
Validation and rollout metrics are calculated on a validation set. The rollouts in Figures \ref{fig:sampling}, \ref{fig:sampling_prepend_vs_no_prepend}, \ref{fig:sampling_multi} and \ref{fig:sampling_diffusion} are sampled from frames of a test set.

\section{Jafar's Batched Masking Logic}
\label{sec:jafar-batched-masking}
\begin{figure*} 
    \centering 
    \begin{Verbatim}[commandchars=\\\{\}]
\PYG{n}{mask\PYGZus{}prob} \PYG{o}{=} \PYG{n}{jax}\PYG{o}{.}\PYG{n}{random}\PYG{o}{.}\PYG{n}{uniform}\PYG{p}{(}\PYG{n}{rng1}\PYG{p}{,} \PYG{n}{minval}\PYG{o}{=}\PYG{n+nb}{self}\PYG{o}{.}\PYG{n}{mask\PYGZus{}limit}\PYG{p}{)}
\PYG{n}{mask} \PYG{o}{=} \PYG{n}{jax}\PYG{o}{.}\PYG{n}{random}\PYG{o}{.}\PYG{n}{bernoulli}\PYG{p}{(}\PYG{n}{rng2}\PYG{p}{,} \PYG{n}{mask\PYGZus{}prob}\PYG{p}{,} \PYG{n}{vid\PYGZus{}embed}\PYG{o}{.}\PYG{n}{shape}\PYG{p}{[}\PYG{o}{:}\PYG{o}{\PYGZhy{}}\PYG{l+m+mi}{1}\PYG{p}{]}\PYG{p}{)}
\PYG{n}{mask} \PYG{o}{=} \PYG{n}{mask}\PYG{o}{.}\PYG{n}{at}\PYG{p}{[}\PYG{p}{:}\PYG{p}{,} \PYG{l+m+mi}{0}\PYG{p}{]}\PYG{o}{.}\PYG{n}{set}\PYG{p}{(}\PYG{k+kc}{False}\PYG{p}{)}
\PYG{n}{vid\PYGZus{}embed} \PYG{o}{=} \PYG{n}{jnp}\PYG{o}{.}\PYG{n}{where}\PYG{p}{(}\PYG{n}{jnp}\PYG{o}{.}\PYG{n}{expand\PYGZus{}dims}\PYG{p}{(}\PYG{n}{mask}\PYG{p}{,} \PYG{o}{\PYGZhy{}}\PYG{l+m+mi}{1}\PYG{p}{),} \PYG{n+nb}{self}\PYG{o}{.}\PYG{n}{mask\PYGZus{}token}\PYG{p}{,} \PYG{n}{vid\PYGZus{}embed}\PYG{p}{)}
\end{Verbatim} 
\caption{Code snippet from \citet{willi2024jafar} showing their batched masking logic.} \label{fig:jafar-batched-masking} 
\end{figure*}

\begin{figure*} 
    \centering 
    \begin{Verbatim}[commandchars=\\\{\}]
\PYG{n}{mask\PYGZus{}prob} \PYG{o}{=} \PYG{n}{jax}\PYG{o}{.}\PYG{n}{random}\PYG{o}{.}\PYG{n}{uniform}\PYG{p}{(}
    \PYG{n}{\PYGZus{}rng\PYGZus{}prob}\PYG{p}{,} \PYG{n}{shape}\PYG{o}{=}\PYG{p}{(}\PYG{n}{batch\PYGZus{}size}\PYG{p}{,}\PYG{p}{),} \PYG{n}{minval}\PYG{o}{=}\PYG{n+nb}{self}\PYG{o}{.}\PYG{n}{mask\PYGZus{}limit}
\PYG{p}{)}
\PYG{n}{per\PYGZus{}sample\PYGZus{}shape} \PYG{o}{=} \PYG{n}{vid\PYGZus{}embed\PYGZus{}BTNM}\PYG{o}{.}\PYG{n}{shape}\PYG{p}{[}\PYG{l+m+mi}{1}\PYG{p}{:}\PYG{o}{\PYGZhy{}}\PYG{l+m+mi}{1}\PYG{p}{]}
\PYG{n}{mask} \PYG{o}{=} \PYG{n}{jax}\PYG{o}{.}\PYG{n}{vmap}\PYG{p}{(}
    \PYG{k}{lambda} \PYG{n}{rng}\PYG{p}{,} \PYG{n}{prob}\PYG{p}{:} \PYG{n}{jax}\PYG{o}{.}\PYG{n}{random}\PYG{o}{.}\PYG{n}{bernoulli}\PYG{p}{(}\PYG{n}{rng}\PYG{p}{,} \PYG{n}{prob}\PYG{p}{,} \PYG{n}{per\PYGZus{}sample\PYGZus{}shape}\PYG{p}{),}
    \PYG{n}{in\PYGZus{}axes}\PYG{o}{=}\PYG{p}{(}\PYG{l+m+mi}{0}\PYG{p}{,} \PYG{l+m+mi}{0}\PYG{p}{),}
\PYG{p}{)(}\PYG{n}{jnp}\PYG{o}{.}\PYG{n}{asarray}\PYG{p}{(}\PYG{n}{\PYGZus{}rngs\PYGZus{}mask}\PYG{p}{),} \PYG{n}{mask\PYGZus{}prob}\PYG{p}{)}
\PYG{n}{mask} \PYG{o}{=} \PYG{n}{mask}\PYG{o}{.}\PYG{n}{at}\PYG{p}{[}\PYG{p}{:}\PYG{p}{,} \PYG{l+m+mi}{0}\PYG{p}{]}\PYG{o}{.}\PYG{n}{set}\PYG{p}{(}\PYG{k+kc}{False}\PYG{p}{)}
\PYG{n}{vid\PYGZus{}embed\PYGZus{}BTNM} \PYG{o}{=} \PYG{n}{jnp}\PYG{o}{.}\PYG{n}{where}\PYG{p}{(}
    \PYG{n}{jnp}\PYG{o}{.}\PYG{n}{expand\PYGZus{}dims}\PYG{p}{(}\PYG{n}{mask}\PYG{p}{,} \PYG{o}{\PYGZhy{}}\PYG{l+m+mi}{1}\PYG{p}{),} \PYG{n+nb}{self}\PYG{o}{.}\PYG{n}{mask\PYGZus{}token}\PYG{o}{.}\PYG{n}{value}\PYG{p}{,} \PYG{n}{vid\PYGZus{}embed\PYGZus{}BTNM}
\PYG{p}{)}
\end{Verbatim} 
\caption{Code snippet from Jasmine showing its batched masking logic.} \label{fig:jasmine-batched-masking}
\end{figure*}

Whereas \citet{willi2024jafar} sample a single masking probability and apply the same masking pattern to all samples in a batch (Figure \ref{fig:jafar-batched-masking}), Jasmine samples \texttt{batch\_size} many sampling probabilities and uses per-sequence masking patterns (Figure \ref{fig:jasmine-batched-masking}). This leads to significantly reduced loss variance (Figure \ref{fig:wc_time}), especially in highly distributed settings.

\section{Model Inspection using Treescope}
\begin{figure}
  \centering
  \includegraphics[width=\textwidth]{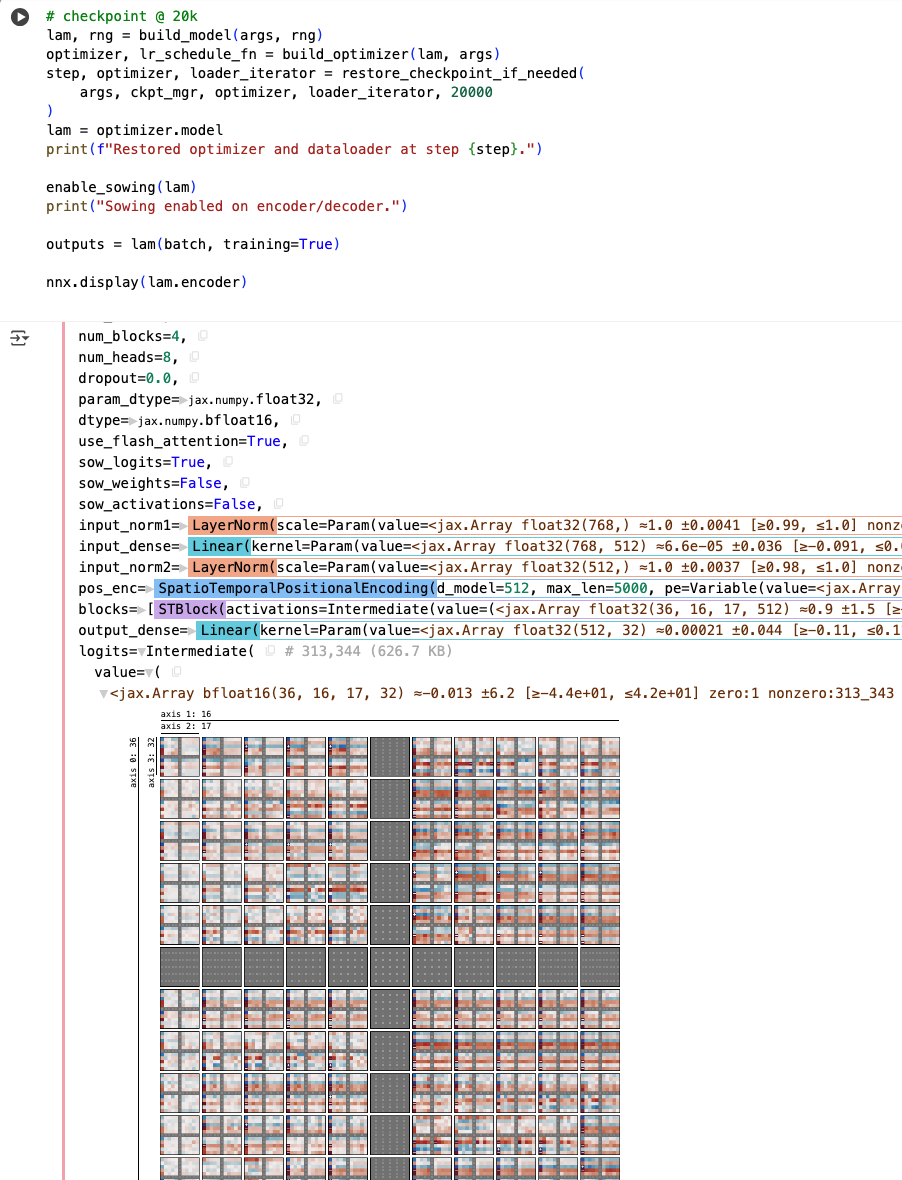}
  \caption{Treescope visualization when performing model inspection. The notebook illustrates inspecting output logits at specific checkpoints.}
  \label{fig:demo-notebook}
\end{figure}

Jasmine's training loop is highly modular and supports easy model surgery as well as model inspection using Treescope \citep{johnson2024penzai}. We provide a demo notebook\footnote{\url{https://colab.research.google.com/drive/1zHkciFIZxXloJgue9F5LtFlA0m00rJIf}} alongside our repository, which illustrates debugging a common training instability (\Cref{fig:demo-notebook}).

\end{document}